\documentclass{article}
\PassOptionsToPackage{numbers, compress}{natbib}
\usepackage[final]{neurips_2022}

\usepackage{amsmath,amssymb} 
\usepackage[utf8]{inputenc} 
\usepackage[T1]{fontenc}    
\usepackage{hyperref}       
\usepackage{url}            
\usepackage{booktabs}       
\usepackage{amsfonts}       
\usepackage{nicefrac}       
\usepackage{microtype}      
\usepackage{xcolor}         
\usepackage{multirow}
\usepackage{graphicx}
\usepackage{algorithm}
\usepackage{algpseudocode}

\title{CAGroup3D: Class-Aware Grouping for 3D Object Detection on Point Clouds}

\author{%
  Haiyang Wang$^{1,6,7}$\footnotemark[1], ~~Lihe Ding$^{2}$\footnotemark[1], ~~Shaocong Dong$^{2}$, ~~Shaoshuai Shi$^{3}$\footnotemark[2], \\
  \textbf{Aoxue Li$^{4}$, ~~Jianan Li$^{2}$, ~~Zhenguo Li$^{4}$, ~~Liwei Wang$^{1,5}$\footnotemark[2]} \\
  {$^1$}Center for Data Science, Peking University ~~{$^2$}Beijing institute of Technology \\
  {$^3$}Max Planck Institute for Informatics ~~{$^4$}Huawei Noah’s Ark Lab, China\\
  {$^5$}Key Laboratory of Machine Perception, MOE, School of Intelligence Science and Technology, \\
  Peking University ~~{$^6$}Peng Cheng Laboratory ~~{$^7$}Pazhou Laboratory (Huangpu) \\
  {\tt\small\{wanghaiyang@stu, wanglw@cls\}.pku.edu.cn, \{dean.dinglihe, shaocong\}@bit.edu.cn}\\
  {\tt\small sshi@mpi-inf.mpg.de, lijianan15@gmail.com \{liaoxue2, Li.Zhenguo\}@huawei.com}\\
}

\begin{document}

\maketitle
\renewcommand{\thefootnote}{\fnsymbol{footnote}}
\footnotetext[1]{Equal contribution.}
\footnotetext[2]{Corresponding author.}
\begin{abstract}
We present a novel two-stage fully sparse convolutional 3D object detection framework, named CAGroup3D. Our proposed method first generates some high-quality 3D proposals by leveraging the class-aware local group strategy on the object surface voxels with the same semantic predictions, which considers semantic consistency and diverse locality abandoned in previous bottom-up approaches. Then, to recover the features of missed voxels due to incorrect voxel-wise segmentation, we build a fully sparse convolutional RoI pooling module to directly aggregate fine-grained spatial information from backbone for further proposal refinement. It is memory-and-computation efficient and can better encode the geometry-specific features of each 3D proposal. Our model achieves state-of-the-art 3D detection performance with remarkable gains of +\textit{3.6\%} on ScanNet V2 and +\textit{2.6}\%  on SUN RGB-D in term of mAP@0.25. Code will be available at \url{https://github.com/Haiyang-W/CAGroup3D}.
\end{abstract}
\section{Introduction}
As a crucial step towards understanding 3D visual world, 3D object detection aims to estimate the oriented 3D bounding boxes and semantic labels of objects in real 3D scenes. It has been studied for a long time in both academia and industry 
since it benefits various downstream applications, such as autonomous driving~\cite{bansal2018chauffeurnet,wang2019monocular}, robotics~\cite{zhu2017target,wang2021collaborative} and augmented reality~\cite{azuma1997survey,billinghurst2015survey}. In this paper, we focus on detecting 3D objects from unordered, sparse and irregular point clouds. Those natural characteristics make it more challenging to directly extend well-studied 2D techniques to 3D detection.

\begin{figure}
  \centering
  \includegraphics[width=\linewidth]{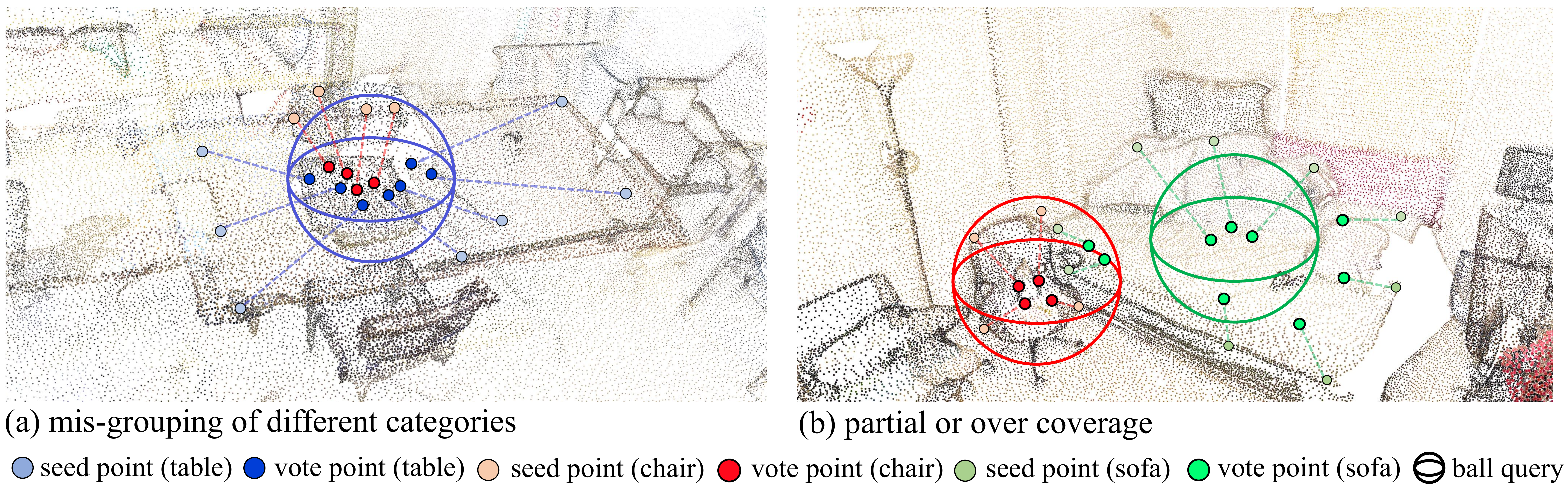}
  \caption{Class-agnostic grouping methods suffer from (a) mis-grouping of different categories within the same local regions, (b) partial coverage of the object surfaces; outliers from the cluttered scene.}
  \label{fig:intro}
\end{figure}

Unlike 3D object detection from autonomous driving scenarios that only considers bird's eye view (BEV) boxes~\cite{shi2019pointrcnn,yin2021cvpr,lang2019pointpillars,fan2021embracing,shi2020p2,shi2020pv}, most of existing 3D indoor object detectors~\cite{qi2019deep,wang2022rbgnet,liu2021group,cheng2021back,zhang2020h3dnet} typically handle this task 
through a bottom-up scheme, which extracts the point-wise features from input point clouds, and then groups the points into their respective instances to generate a set of proposals. 
However, the above grouping algorithms are usually carried out in a class-agnostic manner, which abandons semantic consistency within the same group and also ignores diverse locality among different categories. For example, VoteNet~\cite{qi2019deep} learns the point-wise center offsets and aggregates the points that vote to similar semantic-irrelevant local region. Though impressive, as shown in Figure \ref{fig:intro}, these methods may fail in cluttered indoor scenes where various objects are close but belong to different categories. Also, the object sizes are diverse for different categories, so that a class-agnostic local grouping may partially cover the boundary points of large objects and involve more noise outliers for small objects. 

Hence, we propose CAGroup3D, a two-stage fully convolutional 3D object detection framework. 
Our method consists of two novel components. One is the class-aware 3D proposal generation module, which aims to generate reliable proposals by utilizing class-specific local group strategy on the object surface voxels with same semantic predictions. The other one is an efficient fully sparse convolutional RoI pooling module for recovering the features of the missed surface voxels due to semantic segmentation errors, so as to improve the quality of predicted boxes.  

Specifically, a backbone network with 3D sparse convolution is firstly utilized to extract descriptive voxel-wise features from raw point clouds. Based on the learned features, we conduct a class-aware local grouping module to cluster surface voxels into their corresponding instance centroids. Different from \cite{qi2019deep}, in order to consider the semantic consistency, we not only shift voxels of the same instance towards the same centroid but also predict per-voxel semantic scores. Given the contiguously distributed vote points with their semantic predictions, we initially voxelize them according to the predicted semantic categories and vote coordinates, so as to generate class-specific 3D voxels for different categories. The voxel size of each category is adaptive to its average spatial dimension.
To maintain the structure of fully convolution, we apply sparse convolution as grouping operation centered on each voted voxel to aggregate adjacent voxel features in the same semantic space. Note that these grouping layers are class-dependent but share the same kernel size, thus the larger classes are preferred to be aggregated with larger local regions. 

Secondly, given the proposal candidates, fine-grained specific features within 3D proposals need to be revisited from 3D backbone through certain pooling operation for the following box refinement. 
However, state-of-the-art pooling strategies~\cite{shi2020pv,deng2020voxel} are memory-and-computation intensive due to the hand-crafted set abstraction~\cite{qi2017pointnet}. Besides that, its max-pooling operation also harms the geometry distribution.
To tackle this problem, we propose RoI-Conv pooling module, which directly adopts the well-optimized 3D sparse convolutions to aggregate voxel features from backbone.  It can encode effective geometric representations with a memory-efficient design for further proposal refinement.

In summary, our contributions are three-fold: 
1) We propose a novel class-aware 3D proposal generation strategy, which considers both the voxel-wise semantic consistency within the same local group and the object-level shape diversity among different categories.
2) We present RoI-Conv pooling module, an efficient fully convolutional 3D pooling operation for revisiting voxel features directly from backbone to refine 3D proposals. 
3) Our approach outperforms state-of-the-art methods with remarkable gains on two challenging indoor datasets, \textit{i.e.}, ScanNet V2~\cite{dai2017scannet} and SUN RGB-D~\cite{sunrgbd}, demonstrating its effectiveness and generality.

\section{Related Work}
\textbf{3D Object Detection on Point Clouds.} Detecting 3D objects from point clouds is challenging due to orderless, sparse and irregular characteristics. Previous approaches can be coarsely classified into two lines in terms of point representations, \textit{i.e.}, the voxel-based methods~\cite{zhou2018voxelnet,yang2018pixor,shi2020p2,yan2018second,shi2020pv,yin2021cvpr,shi2020p2} and the point-based methods~\cite{qi2019deep,wang2022rbgnet,cheng2021back,liu2021group,zhang2020h3dnet,yang2022boosting}.  Voxel-based methods are mainly applied in outdoor autonomous driving scenarios where objects are distributed on the large-scale 2D ground plane. They process the sparse point clouds by efficient 3D sparse convolution, then project these 3D volumes to 2D grids for detecting bird's eye view (BEV) bboxes by 2D ConvNet. Powered by PointNet series~\cite{qi2017pointnet,qi2018pointnnetplus}, point-based methods are also widely used to predict 3D bounding bboxes. Most of existing methods are in a bottom-up manner, which extracts the point-wise features and groups them to obtain object features. This pipeline has been a great success for estimating 3D bboxes directly from cluttered and dense 3D scenes. However, due to the hand-crafted point sampling and computation intensive grouping scheme applied in PointNet++~\cite{qi2018pointnnetplus}, they are difficult to be extended to large-scale point clouds. Hence, we propose an efficient fully convolutional bottom-up framework to efficiently detect 3D bboxes directly from dense 3D point clouds.

\textbf{Feature Grouping.} Feature grouping is a crucial step for bottom-up 3D object detectors~\cite{qi2019deep,wang2022rbgnet,liu2021group,cheng2021back,zhang2020h3dnet,vu2022softgroup}, which clusters a group of point-wise features to generate high-quality 3D bounding boxes. Among the numerous successors, voting-based framework~\cite{qi2019deep} is widely used, which groups the points that vote to the same local region. Though impressive, it doesn't consider the semantic consistency, so that may fail in cluttered indoor scenes where the objects of different classes are distributed closely. Moreover, voting-based methods usually adopt a class-agonistic local region for all objects, which may incorrectly group the boundary points of large objects and involve more noise points for small objects. To address the above limitations, we present a class-aware local grouping strategy to aggregate the points of the same category with class-specific center regions.

\textbf{Two-stage 3D Object Detection.} Many state-of-the-art methods considered applying RCNN style 2D detectors to the 3D scenes, which apply 3D RoI-pooling scheme or its variants~\cite{shi2019pointrcnn,shi2020p2,deng2020voxel,shi2020pv,yang2019std,xu2022fusionrcnn} to aggregate the specific features within 3D proposals for the box refinement in a second stage. These pooling algorithms are usually equipped with set abstraction~\cite{qi2017pointnet} to encode local spatial features, which consists of a hand-crafted query operation (\textit{e.g.}, ball query~\cite{qi2017pointnet} or vector query~\cite{deng2020voxel}) to capture the local points and a max-pooling operation to group the assigned features. Therefore these RoI pooling modules are mostly computation expensive. Moreover, the max-pooling operation also harms the spatial distribution information. To tackle these problems, we propose RoI-Conv pooling, a memory-and-computation efficient fully convolutional RoI pooling operation to aggregate the specific features for the following refinement. 

\section{Methodology}
In this paper, we propose CAGroup3D, a two-stage fully convolutional 3D object detection framework for estimating accurate 3D bounding boxes from point clouds. The overall architecture of CAGroup3D is depicted in Figure \ref{fig:overview}. Our framework consists of three major components: an efficient 3D voxel CNN with sparse convolution as the backbone network for point cloud feature learning (\S \ref{sec:backbone}), a class-aware 3D proposal generation module for predicting high quality 3D proposals by aggregating voxel features of the same category within the class-specific local regions (\S \ref{sec:proposal}) and RoI-Conv pooling module for directly extracting complete and fine-grained voxel features from the backbone 
to revisit the miss-segmented surface voxels and refine 3D proposals. Finally, we formulate the learning objective of our framework in \S \ref{sec:loss}. 
 
\subsection{3D Voxel CNN for Point Cloud Feature Learning}\label{sec:backbone}
For generating accurate 3D proposals, we first need to learn discriminative geometric representation for describing input point clouds. Voxel CNN with 3D sparse convolution~\cite{shi2020p2,yan2018second,zhou2018voxelnet,graham20183d,graham2017submanifold} is widely used by state-of-the-art 3D detectors thanks to its high efficiency and scalability of converting the point clouds to regular 3D volumes. In this paper, we adopt sparse convolution based backbone for feature encoding and 3D proposal generation.

\begin{figure}
  \centering
  \includegraphics[width=1.0\linewidth]{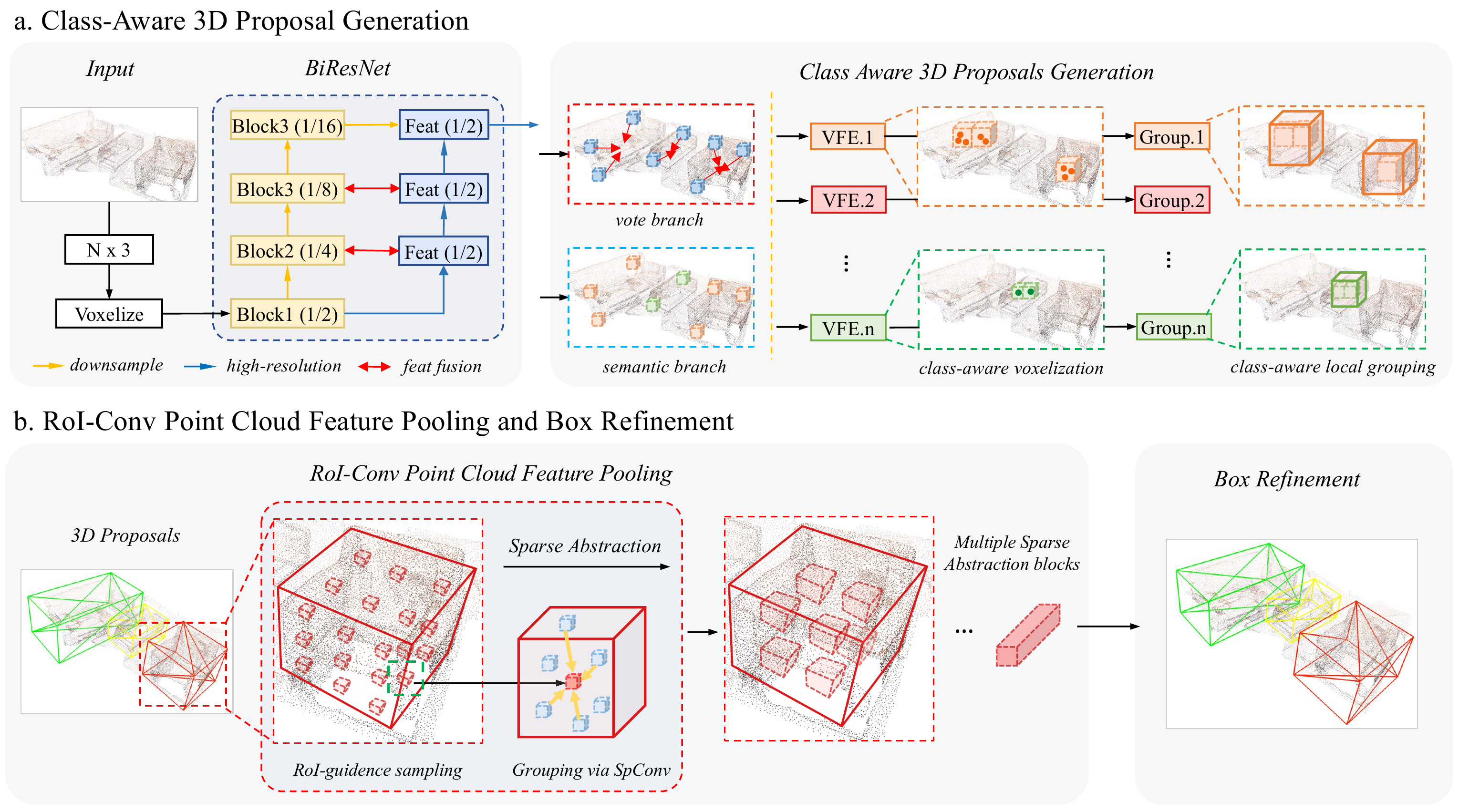}
  \caption{The overall architecture of CAGroup3D. (a) Generate 3D proposals by utilizing class-aware local grouping on the vote space with same semantic predictions. (b) Aggregating the specific features within the 3D proposals by the efficient RoI-Conv pooling module for the following box refinement.}
  \label{fig:overview}
\end{figure}

3D backbone network equipped with high-resolution feature maps and large receptive fields is critical for accurate 3D bounding box estimation and voxel-wise semantic segmentation. The latter is 
closely related to the accuracy of succeeding grouping module. To maintain these two characteristics, inspired by the success of HRNet series~\cite{wang2020deep,hong2021deep,sun2019deep} in segmentation community, we implement a 3D voxel bilateral network with dual resolution based on ResNet~\cite{he2016deep}. For brevity, we refer it as BiResNet. As shown in Figure \ref{fig:overview}, our backbone network contains two branches. One is the sparse modification of ResNet18~\cite{he2016deep} where all 2D convolutions are replaced with 3D sparse convolutions. It can extract multi-scale contextual information 
with proper downsampling modules.
The other one is a auxiliary branch that maintains a high-resolution feature map whose resolution is 1/2 of the input 3D voxels. Specifically, the auxiliary branch is inserted following the first stage of ResNet backbone and doesn't contain any downsampling operation. Similar to \cite{wang2020deep}, we adopt the bridge operation between the two paths to perform the bilateral feature fusion. Finally, the fine-grained voxel-wise geometric features with rich contextual information are generated by the high-resolution branch and facilitate the following module. Experiments also demonstrate that our voxel backbone performs better than previous FPN-based ResNet~\cite{lin2017feature}. More architecture details are in Appendix.

\subsection{Class-Aware 3D Proposal Generation}\label{sec:proposal}
Given the voxel-wise geometric features generated by the backbone network, a bottom-up grouping algorithm 
is generally adopted
to aggregate object surface voxels into their respective ground truth instances and generate reliable 3D proposals. Voting-based grouping method~\cite{qi2019deep} has shown great success for 3D object detection, which is performed in a class-agnostic manner.
It reformulates Hough voting to learn point-wise center offsets, and then generates object candidates by clustering the points that vote to similar center regions.
However, this method may incorrectly group the outliers in the cluttered indoor scenarios (\textit{e.g.}, votes are close together but belong to different categories), which 
degrades the performance of 
3D object detection. Moreover, due to the diverse object sizes of different categories, class-agnostic local regions may mis-group the boundary points of large objects and involve more noise points for small objects.

To address this limitation, we propose the class-aware 3D proposal generation module, which first produces voxel-wise predictions (\textit{e.g.}, semantic maps and geometric shifts), and then clusters the object surface voxels of the same semantic predictions with class-specific local groups. 

\textbf{Voxel-wise Semantic and Vote Prediction.}
After obtaining the voxel features from backbone network, two branches are constructed to output the voxel-wise semantic scores and center offset vectors. Specifically, the backbone network generates a number of $N$ non-empty voxels $\{o_i\}^{N}_{i=1}$ from backbone, where $o_i = [x_i;f_i]$ with $x_i \in \mathbb{R}^3$ and $f_i \in \mathbb{R}^C$. A voting branch encodes the voxel feature $f_i$ to learn the spatial center offset $\Delta x_i\in\mathbb{R}^3$ and feature offset $\Delta f_i\in\mathbb{R}^C$. Based on the learned spatial and feature offset, we shift voxel $o_i$ to the center of its respective instance and generate vote point $p_i$ as follow:
\begin{equation}
		\left\{p_i\mid p_i=[x_i+\Delta x_i, f_i+\Delta f_i] \right\}_{i=1}^N.
\end{equation}
The predicted offset $\Delta x_i$ is explicitly supervised by a smooth-$\ell_1$ loss with the ground-truth displacement from the coordinate of seed voxel $x_i$ to its corresponding bounding box center.

In parallel with the voting branch, we also construct a semantic branch to output semantic scores $S=\{s_i\}^{N}_{i=1}$ for all the voxels over $N_{class}$ classes as
\begin{equation}
    \begin{aligned}
    s_i = \text{MLP}^{\text{sem}}(o_{i}) \in [0,1]^{N_{class}}, ~~~\text{for}~ i=1, \cdot\cdot\cdot,N, 
    \end{aligned}
\end{equation}
where $\text{MLP}^{\text{sem}}(\cdot)$ is a one-layer multi-layer-perceptron (MLP) network and $s_i$ indicates the semantic probability for all classes of voxel $o_i$. We adopt focal loss~\cite{lin2017focal} for calculating voxel segmentation loss to handle the class imbalance issue.

Notably, the vote and semantic targets of each voxel are associated with the ground-truth 3D boxes not the instance or semantic masks, so that it can be easily generalized to the 3D object detection datasets with bounding box annotations. To be specific, for each voxel, only the ground-truth bounding boxes that includes this voxel are selected. Considering the ambiguous cases that a voxel is in multiple ground truth bounding boxes, only the box with the least volume is assigned to this voxel. 

\textbf{Class-Aware Local Grouping.} This step aims to produce reliable 3D proposals in a bottom-up scheme based on the above voxel-wise semantic scores $\{s_i\}^{N}_{i=1}$ and vote predictions $\{p_i\}^{N}_{i=1}$. To carry out grouping with semantic predictions, we first define a score threshold $\tau$ for all  categories to 
individually determine whether a voxel belongs to a category instead of utilizing the one-hot semantic predictions.
It can allow the voxel to be associated with multiple classes and thus improve the recall of semantic voxels for each category. Given the semantic scores $S=\{s_i\}^{N}_{i=1}$, we iterate over all the classes, and slice a point subset from the whole scene of each class that has the score higher than the threshold $\tau$, to form the class-dependent vote set $\{c_j\}_{j=1}^{N_{class}}$,
\begin{equation}
	\left\{c_j\mid c_j=\{p_i: s_i^j > \tau , i=1, ..., N \} \right\}_{j=1}^{N_{class}}.
	\label{eq:vfe}
\end{equation}
The above semantic subset is generated in a contiguous euclidean space and the vote points are distributed irregularly. 
To maintain the structure of pure convolutions and facilitate the succeeding class-aware convolution-based local grouping module, we individually voxelize the vote points in each semantic subset to $\{V_j\}_{j=1}^{N_{class}}$  by employing a voxel feature encoding (VFE) layer with a class-specific 3D voxel size, which is proportional to the class average spatial dimension. Specifically, this re-voxelization process for each class can be formulated as follows:
\begin{equation}
	\{v_i\}_{i=1}^{\mid V_j \mid} =\text{VFE}(c_j, ~\alpha \cdot d_j, ~Avg), ~~~\text{for}~ j = 1, ..., N_{class},
\end{equation}
where $c_j$ is the class-dependent vote set of class index $j$ and $\mid V_j \mid$ is the number of non-empty voxels after class individual re-voxelization. $\alpha \cdot d_j$ is the class-specific voxel size, where $\alpha$ is a predefined scale factor and $d_j = (w_j, h_j, l_j)$ is the category average spatial dimension. $\text{VFE}(~\cdot~, ~\alpha \cdot d_j, ~Avg)$ means that the average pooling operation is adopted to voxelize vote features on $j\text{-th}$ class subset with the voxel size $\alpha \cdot d_j$. Importantly, the voxel size is adaptive among different categories, which is more diverse than the widely used FPN-based prediction structure.

Given the predicted vote voxels of $j\text{-th}$ class $\{v_i\}_{i=1}^{\mid V_j \mid}$, we apply sparse convolutions with a predefined kernel size $k^{(a)}$ on each voxel, and automatically aggregate local context inside the class to generate class-specific geometric features $A^{(j)}$ as follow:
\begin{equation}
    A^{(j)} = \{a_i^{(j)} \mid a_i^{(j)} = \text{SparseConv}_{\text{3D}}^{(j)}(v_i, ~ \{v_i\}_{i=1}^{\mid V_j \mid}, ~k^{(a)})\}_{i=1}^{\mid V_j \mid},
    \label{eq:multiconv}
\end{equation}
where $\text{SparseConv}^{(j)}_{\text{3D}}(\cdot_{\text{ center}}, \cdot_{\text{ support voxels}}, \cdot_{\text{ kernel size}})$ is the standard sparse 3D convolution~\cite{graham20183d} and specific for different classes. 
A shared anchor-free head is appended to the aggregated features for predicting classification probabilities, bounding box regression parameters and confidence scores.

\subsection{RoI-Conv point cloud feature pooling for 3D Proposal Refinement}\label{sec:refine}
Due to the semantic segmentation errors in stage-I, class-aware local grouping module will mis-group some object surface voxels. So an efficient pooling module is needed to recover the missed voxel features and also aggregate more fine-grained features from backbone for proposal refinement. 
State-of-the-art 3D RoI pooling strategies~\cite{shi2020pv,deng2020voxel} usually adopt set abstraction operation to encode local patterns, which is computation expensive compared to traditional convolution and hand-crafted with lots of hyper-parameters (\textit{e.g.}, radius, the number of neighbors). Moreover, its max-pooling operation also harms the spatial distribution information. 

To tackle these limitations and hold a fully convolution structure, we propose RoI-Conv pooling operation, which builds a hierarchical grouping module with well optimized 3D sparse convolutions to directly aggregate RoI-specific features from backbone for further proposal refinement. Our hierarchical structure is composed by a number of \textit{sparse abstraction} block, which contains two key components: the RoI-guidence sampling for selecting a subset of input voxels within the 3D proposals, which defines the centroids of local regions, and a shared sparse convolution layer for encoding local patterns into feature vectors. 

\textbf{Sparse Abstraction.}
As shown in Figure \ref{fig:overview}, the inputs to this block are a number of $\mid \mathcal{I} \mid$ input voxels  $\mathcal{I} = \{l_{n}\}_{n=1}^{\mathcal{I}}$ and $\mid \mathcal{M} \mid$ proposals $\mathcal{M} = \{\rho_m\}_{m=1}^{\mathcal{M}}$, where $\rho_m$ is the proposal parameters for guiding the point sampling. The output is a number of $\mid \mathcal{Q} \mid$ pooling RoI-specific voxels $\mathcal{Q} = \{q_{k}\}_{k=1}^{\mid \mathcal{Q} \mid}$.

Specifically, given the input voxels and proposals, instead of directly sampling from the whole input space, we adopt the RoI-guidence sampling to uniformly sample $G_x \times G_y \times G_z$ grid points within each 3D proposal in voxel space, which are denoted as $\mathcal{G}=\{g_k \in \mathbb{Z}^3\}^{G_x \times G_y \times G_z \times \mathcal{M}}_{k=1}$. $G_x \times G_y \times G_z$ is the proposal sampling resolution, which are the hyper-parameters independent on proposal sizes. Considering the overlap of different proposals, we merge the repeated grid points and generate a unique points set $\widetilde{\mathcal{G}}=\{g_k\}^{\mid \widetilde{\mathcal{G}} \mid}_{k=1}$, where $\mid \widetilde{\mathcal{G}} \mid$ is the number of unique grid points. Then, with the RoI-specific points set $\widetilde{\mathcal{G}}$, we exploit a sparse convolution centered on each sampled point to cover a set of neighboring input voxels (\textit{e.g.}, $\mathcal{N}_k = \{l_k^1, l_k^2, ..., l_k^{L_k}\}$ for $g_k$) within the kernel size $k^{(p)}$ as:
\begin{equation}
    \mathcal{Q} = \left\{\begin{aligned}
    q_k \mid q_k = \left\{
    		\begin{aligned}
            		&\text{SparseConv}_{\text{3D}}(g_k, ~ \mathcal{N}_k, ~k^{(p)}),~~~\text{if}  ~~L_k  > 0,~~~~~\\ 
            		&\phi,~~~~~~~~~~~~~~~~~~~~~~~~~~~~~~~~~~~~~~~~~~~~~~~\text{if} ~~L_k  = 0,~~~~~\\ 
    		\end{aligned}
    		\right.\!\!
    	\end{aligned}
    	\text{for}~ k = 1, ..., \mid \widetilde{\mathcal{G}} \mid
    	\right \}
    		\label{eq:griconv}
\end{equation} 
where $L_k$ is the number of neighboring voxels queried by the $k\text{-th}$ RoI-specific point with kernel size $k^{(p)}$. $\text{SparseConv}_{\text{3D}}(\cdot_{\text{ center}}, \cdot_{\text{ support voxels}}, \cdot_{\text{ kernel size}})$ is a shared sparse 3D convolution for all the proposals and only applied on the points that their neighboring voxel sets are non-empty. $\phi$ means empty voxel. Then, to hold the surface geometry and reduce computation cost, we abandon the empty voxels and output the RoI-specific voxels set $\mathcal{Q}$.

\textbf{RoI-Conv Pooling Module.} Our pooling network is equipped with two-layers sparse abstraction block and progressively abstracts the voxel features within each 3D proposal from the backbone to RoI-specific features iteratively as:
\begin{equation}
    \begin{aligned}
    & \mathcal{Q}^{1} =  \text{SparseAbs}(\mathcal{I}, ~  \{\rho_m\}_{m=1}^{\mathcal{M}}, ~ 7 \times 7 \times 7, ~5), ~~~~~\\
    & \mathcal{F} = \text{SparseAbs}( \mathcal{Q}^{1}, ~  \{\rho_m\}_{m=1}^{\mathcal{M}}, ~1 \times 1 \times 1, ~ 7),
    \end{aligned}
\end{equation}
where $\text{SparseAbs}(\cdot_{\text{ input voxels}}, \cdot_{\text{ proposals}}, \cdot_{\text{sampling resolution}}, \cdot_{\text{ kernel size}})$ denotes our sparse abstraction block.
Notably, to encode the voxel features of oriented proposals, we follow the transformation strategy in \cite{shi2020p2} before the last block, and normalize the input voxels belonging to each proposal to its individual canonical systems. With the RoI feature $\mathcal{F} \in R^C$ of each proposal, the refinement network predicts the size and location (\textit{i.e.}, bbox dimension, center and orientation) residuals relative to the proposal in Stage-I and the targets are encoded by the traditional residual-based method ~\cite{shi2020pv,shi2020p2}.

\subsection{Learning Objective}\label{sec:loss}
Our proposed approach is trained from scratch with semantic loss $\mathcal{L}_\text{sem}$, voting loss $\mathcal{L}_{\text{vote-reg}}$, centerness loss $\mathcal{L}_\text{cntr}$, bounding box estimation loss $\mathcal{L}_\text{box}$, classification losses $\mathcal{L}_\text{cls}$ for Stage-I and bbox refinement loss $\mathcal{L}_\text{rebox}$ for Stage-II, which are formulated as follows:

\begin{equation}
    \begin{aligned}
    L = \beta_\text{sem}\mathcal{L}_\text{sem} + \beta_\text{vote}\mathcal{L}_{\text{vote}} + \beta_\text{cntr}\mathcal{L}_\text{cntr} \\ 
    +\beta_\text{box}\mathcal{L}_\text{box} + \beta_\text{cls}\mathcal{L}_\text{cls} + \beta_\text{rebox}\mathcal{L}_\text{rebox}.
    \end{aligned}
\end{equation}
$\mathcal{L}_\text{sem-cls}$ is a multi-class focal loss~\cite{lin2017focal} used to supervise voxel-wise semantic segmentation. $\mathcal{L}_{\text{vote}}$ is a smooth-$\ell_1$ loss for predicting the center offset of each voxel. In term of the 3D proposal generation module, we follow the same loss functions $\mathcal{L}_\text{cntr}$, $\mathcal{L}_\text{box}$ and $\mathcal{L}_\text{cls}$ defined in \cite{rukhovich2021fcaf3d} to optimize object centerness, bounding box estimation and classification respectively. For the second stage, $\mathcal{L}_\text{rebox}$  is the residual-based smooth-$\ell_1$ box regression loss for 3D box proposal refinement, which contains size, box center and angle refinement loss. Besides that, we also add the same IoU loss $\mathcal{L}_\text{iou}$ as used in stage-I, and the final box refinement loss is as follows

\begin{equation}
    \mathcal{L}_\text{rebox} = \sum\limits_{r\in\{x,y,z,l,h,w,\theta \}} \mathcal{L}_{\text{smooth-}\ell_1}(\Delta r^*, \Delta r) + \mathcal{L}_{\text{iou}},
\end{equation}
where $\Delta r$ is the predicted residual and $\Delta r^*$ is the corresponding ground truth. The detailed balancing factors are in Appendix.

\section{Experiments}

\begin{table}
  \caption{3D detection results on ScanNet V2~\cite{dai2017scannet} and SUN RGB-D~\cite{sunrgbd}. The main comparison is based on the best results of multiple experiments, and the average value of 25 trials is given in brackets. * means the multi-sensor approaches that use both point clouds and RGB images.}
  \label{tab:results}
  \centering
  \begin{tabular}{cccccc}
    \toprule
    & & \multicolumn{2}{c}{ScanNet V2} & \multicolumn{2}{c}{SUN RGB-D} \\
    \cmidrule(r){3-6}
    \multirow{-2}{*}{Methods}     & \multirow{-2}{*}{Presened at}    & mAP@0.25 & mAP@0.5 & mAP@0.25 & mAP@0.5\\
    \midrule
    F-PointNet~\cite{qi2018frustum}*   			& CVPR'18  & 19.8 & 10.8 & 54.0 & -   \\
	GSPN~\cite{yi2018gspn}*                     & CVPR'19  & 30.6 & 17.7 & -    & -   \\
	3D-SIS~\cite{hou20193d}*       				& CVPR'19  & 40.2 & 22.5 & -    & -   \\
	ImVoteNet~\cite{qi2020imvotenet}*       		& CVPR'20  & -    & -    & 63.4 & -   \\
	TokenFusion~\cite{wang2022multimodal}*      &CVPR'22   & 70.8(69.8) & 54.2(53.6) & 64.9(64.4) & 48.3(47.7)\\
	\midrule
	VoteNet~\cite{qi2019deep}                   & ICCV'19  & 58.6 & 33.5 & 57.7 & -   \\
	3D-MPA~\cite{engelmann20203d}       			& CVPR'20  & 64.2 & 49.2 & -    & -   \\
	HGNet~\cite{chen2020hierarchical}        	& CVPR'20  & 61.3 & 34.4 & 61.6 & -   \\
	MLCVNet~\cite{xie2020mlcvnet}               & CVPR'20  & 64.5 & 41.4 & 59.8 & -   \\
	GSDN~\cite{gwak2020generative}              & ECCV'20  & 62.8 & 34.8 & -    & -   \\
	H3DNet~\cite{zhang2020h3dnet}       			& ECCV'20  & 67.2 & 48.1 & 60.1 & 39.0\\
	BRNet~\cite{cheng2021back}        			& CVPR'21  & 66.1 & 50.9 & 61.1 & 43.7\\
	3DETR~\cite{misra2021end}        			& ICCV'21  & 65.0 & 47.0 & 59.1 & 32.7\\
	VENet~\cite{xie2021venet}        			& ICCV'21  & 67.7 & -    & 62.5 & 39.2\\
	Group-free~\cite{liu2021group}   			& ICCV'21  & 69.1(68.6) & 52.8(51.8) & 63.0(62.6) & 45.2(44.4)\\
	RBGNet~\cite{wang2022rbgnet}   			    & CVPR'22  & 70.6(69.6) & 55.2(54.7) & 64.1(63.6) & 47.2(46.3)\\
	HyperDet3D~\cite{zheng2022hyperdet3d}       & CVPR'22  & 70.9       & 57.2       & 63.5       & 47.3     \\
	FCAF3D~\cite{rukhovich2021fcaf3d} 			& ECCV'22	   & 71.5(70.7) & 57.3(56.0) & 64.2(63.8) & 48.9(48.2)\\
	Ours			                            & - 	       & \textbf{75.1(74.5)} & \textbf{61.3(60.3)} & \textbf{66.8(66.4)} & \textbf{50.2(49.5)} \\
    \bottomrule
  \end{tabular}
\end{table}

\subsection{Datasets and Evaluation Metric}\label{sec:de}
Our CAGroup3D is evaluated on two indoor challenging 3D scene datasets, \textit{i.e.}, ScanNet V2~\cite{dai2017scannet} and SUN RGB-D~\cite{sunrgbd}. For all datasets, we follow the standard data splits adopted in \cite{qi2019deep}.

\textbf{ScanNet V2} contains richly-annotated 3D reconstructed indoor scenes with axis-aligned bounding box for most common 18 object categories. It contains 1201 training samples and the remaining 312 scans are left for validation.  We follow \cite{qi2019deep} to sample point clouds from the reconstructed meshes.

\textbf{SUN-RGB-D} is a single-view indoor dataset which consists of 10,355 RGB-D images for 3D scene understanding. It contains $\sim$5K training images annotated with the oriented 3D bounding boxes and the semantic labels for 10 categories. To feed the point data to our method, we follow \cite{qi2019deep} and convert the depth images to point clouds using the provided camera parameters.
 
All the experiment results on both datasets are evaluated by a standard evaluation protocol~\cite{qi2019deep,rukhovich2021fcaf3d}, which uses mean average precision(mAP) with different IoU thresholds, \textit{i.e.}, 0.25 and 0.50.
\subsection{Implementation Details}\label{sec:imple}
\textbf{Network Architecture Details.} For both datasets, we set the voxel size as $0.02m$. As for the backbone, we use the same 3D voxel ResNet18 introduced in \cite{gwak2020generative} as the downsample branch to extract rich contextual information and set the voxel size of high-resolution branch to $0.04m$. In terms of class-aware 3D proposal generation module, we set the scale factor of re-voxelization $\alpha$ to 0.15. The semantic threshold $\tau$ is initially set to 0.15 and then decreases by 0.02 every 10 epochs for ScanNet V2 and 4 epochs for SUN RGB-D until it reaches the minimum value of 0.05. Moreover, the kernel size of convolution-based grouping module $k^{(a)}$ is 9. We stack two sparse abstraction blocks for extracting RoI-specific representations, in where the proposal sampling resolutions are set to $7 \times 7 \times 7$ and $1 \times 1 \times 1$, and the kernel sizes of sparse grouping $k^{(p)}$ are set to $5$ and $7$ respectively.

\begin{table}
  \caption{Effect of Semantic Prediction, Diverse Local Group, RoI-Conv and BiResNet.}
  \label{tab:main_ab}
  \centering
  \begin{tabular}{cccccc}
    \toprule
    Semantic Prediction & Diverse Local Group & BiResNet & RoI-Conv & mAP@0.25 & mAP@0.5\\
    \midrule
    		     	    &                     &            &            & 68.22        & 53.17      \\
    \checkmark   	    &                     &            &            & 69.24        & 54.05      \\
    \checkmark   		& \checkmark          &            &            & 72.10        & 57.07      \\
    \checkmark   		& \checkmark          & \checkmark &            & 73.21        & 57.18      \\
    \checkmark   		& \checkmark          & \checkmark & \checkmark & \textbf{74.50}        & \textbf{60.31}      \\
	
    \bottomrule
  \end{tabular}
\end{table}

\begin{table}[t]
  \begin{minipage}[b]{0.48\linewidth}
  	\caption{Ablation study of the class-aware local group with different scale factor $\alpha$. ($k^{(a)}=9$)}
  	\label{tab:scale_ab}
  	\centering
  	\begin{tabular}{ccc}
    	\toprule
    	scale factor $\alpha$ & mAP@0.25 & mAP@0.5\\
    	\midrule
    	1.00		    & 36.38        & 24.66      \\
    	0.60   	    & 61.49        & 47.41      \\
    	0.20   		& 74.21        & 58.77      \\
    	0.15   		& \textbf{74.50}        & \textbf{60.31}      \\
    	0.10   		& 74.01        & 59.03      \\
    	0.05        & 72.98        & 58.10      \\
	
    	\bottomrule
  	\end{tabular}
  \end{minipage}
  \hspace{12pt}
  \begin{minipage}[b]{0.48\linewidth}
  	\caption{Ablation study of the class-aware local group with different semantic threshold $\tau$.}
  	\label{tab:thres_ab}
  	\centering
  	\begin{tabular}{ccc}
    	\toprule
    	sem thres $\tau$ & mAP@0.25 & mAP@0.5 \\
    	\midrule
    	0.20		     	& 73.51        & 59.50      \\
    	0.10  	            & 74.28        & 59.97      \\
    	0.08  	            & 74.38        & 60.10      \\
    	0.06   				& \textbf{74.51}        & \textbf{60.27}      \\
    	0.04   				& 74.38        & 59.98      \\
    	0.02  				& 73.53        & 58.85      \\
	
    	\bottomrule
  	\end{tabular}
  \end{minipage}
\end{table}

\textbf{Training and Evaluation Scheme.} Our model is trained in an end-to-end manner by AdamW optimizer~\cite{loshchilov2018decoupled}. Following the ~\cite{rukhovich2021fcaf3d}, we set batch size, initial learning rate and weight decay are 16, 0.001 and 0.0001 for both datasets. Training ScanNet V2 requires 120 epochs with the learning rate decay by 10x on 80 epochs and 110 epochs. SUN RGB-D takes 48 epochs and learning rate decayed on 32, 44 epochs. All models are trained on two NVIDIA Tesla V100 GPUs with a 32 GB memory per-card. The gradnorm clip~\cite{zhang2020_b282d173, jin2021nonconvex} is applied to stabilize the training dynamics. We follow the same evaluation scheme in ~\cite{liu2021group}, which runs training for 5 times and test each trained model for 5 times. We report both the best and average metrics across all results. 
\subsection{Benchmarking Results}\label{sec:compare}
We report the comparison results with state-of-the-art 3D detection methods on ScanNet V2~\cite{dai2017scannet}] and SUN RGB-D~\cite{sunrgbd} benchmark. Same as  \cite{liu2021group}, we report the best and average results of 5$\times$5 trials.

As shown in Table \ref{tab:results}, our approach leads to 75.1 in terms of mAP@0.25 and 61.3 in terms of mAP@0.50 on ScanNet V2~\cite{dai2017scannet}, which is +3.6 and +4.0 better than the state-of-the-art~\cite{rukhovich2021fcaf3d}.  For SUN RGB-D, our approaches achieves 66.8 and 50.2, which gains +2.6 and +1.3 in terms of mAP, with 3D IoU threshold 0.25 and 0.5 respectively. Moreover, our approach outperforms all the multi-sensor based methods, which is more challenging for SUN RGB-D due to its relatively poor point clouds. 

\subsection{Ablation Studies and Discussions}\label{sec:ab}
We conduct extensive ablation studys on the \textit{val} sets of ScanNet V2 to analyze individual components of our proposed method. Following \cite{liu2021group}, we report the average performance of 25 trials by default.

\textbf{Effect of class-aware local grouping module.} We first ablate the effects of class-aware local grouping module in Table \ref{tab:main_ab}, \ref{tab:scale_ab}, \ref{tab:thres_ab}. In Table \ref{tab:main_ab}, the base competitor ($1^{st}$ row) is the fully sparse convolutional VoteNet~\cite{qi2019deep} we implemented. Compared with CAGroup3D, it abandons the two stage refinement, replaces the BiResNet with FPN-based variant and groups the vote voxels with the voxel size of 0.02 in a class-agnostic manner. As evidenced in the $1^{st}$ and the $2^{nd}$ rows, by considering the semantic consistency, our model performs better, \textit{i.e.}, 68.22 $\rightarrow$ 69.24, 53.17 $\rightarrow$ 54.05 on mAP@0.25 and mAP@0.5.  Combining the semantic predictions with class-specific local groups ($3^{rd}$ rows), our model achieves a significant improvement, \textit{i.e.}, 68.22 $\rightarrow$ 72.10, 53.17 $\rightarrow$ 57.07. That verifies our motivation that semantic consistency within the same group and diverse locality among different categories are crucial for an effective grouping algorithm. 

Our class-aware local grouping module also works well for a range of hyper-parameters, such as the scale factor $\alpha$ of class-specific re-voxelization and the semantic threshold $\tau$. Table \ref{tab:scale_ab} shows the performance of our model with different scale factors. We first set $k^{(a)}$ to 9 and reduce the scale factor $\alpha$ gradually. With the decrease of $\alpha$, the output voxels are more fine-grained, which is beneficial to estimate accurate bounding boxes and localize small objects. However, a very small voxel size will degrade the performance. With the fixed $k^{(a)}$, smaller voxel sizes will lead to smaller local regions, which may mis-group the boundary object points, and thus it is hard for the network to accurately capture local object geometry. Table \ref{tab:thres_ab} also ablates the effectiveness of different semantic threshold. We observe that the performance gradually improves with the decrease of $\tau$. However, too small threshold can also drops the performance due to the abandonment of semantic consistency.

\textbf{Effect of RoI-Conv pooling module.} Table \ref{tab:main_ab} demonstrates the effectiveness of two-stage refinement with RoI-Conv pooling module. By comparing the $4^{rd}$ and $5^{th}$ rows, we find that our refinement module can really help to detect more accurate 3D bounding boxes, especially in term of mAP@0.50, \textit{i.e.}, 57.18 $\rightarrow$ 60.31. To further ablate the high performance of our RoI-Conv pooling module, we compare it to several pooling strategies~\cite{shi2019pointrcnn,shi2020p2} widely used in 3D object detection. For a fair comparison, we only switch the RoI pooling algorithm while all other settings remain unchanged, (\textit{e.g.}, class-aware local grouping and BiResNet). Table \ref{tab:roi_ab} shows that our approach surpasses others on the performance of both detection scores and computation cost with a remarkable margin. Move details about the above competitors are in Appendix. Table \ref{tab:depth_ab} shows the impact of stacking different number of sparse abstraction blocks with $k^{(p)} \text{=} ~5$ except the last layer. We choose the relative shallow design of two layers as there are no obvious improvement by additional deepening. We further ablate the influence of proposal sampling resolution and sparse kernel size $k^{(p)}$ based on our two layers architecture. Table \ref{tab:partsize_ab} and \ref{tab:pk_ab} show that both the larger $G$ and $k^{(p)}$ can capture more fine-grained geometric details and lead to better performance. Considering the trade-off between memory usage and performance improvement, our model finally sets $G$ and $k^{(p)}$ of the $1^{st}$ layer to $7 \times 7 \times 7$ and 5. 

\textbf{Effect of bilateral feature learning.} In Table \ref{tab:main_ab}, we investigate the effects of bilateral backbone (BiResNet18) by replacing it with FPN-based ResNet18~\cite{rukhovich2021fcaf3d,gwak2020generative}. The $3^{th}$ and $4^{th}$ rows exhibit that the performance drops, especially in term of mAP@0.25, from 73.21 $\rightarrow$ 72.10, when adopts FPN-based backbone. This phenomenon validates that our bilateral backbone could learn much richer contextual information while maintain the high-resolution representations.

\begin{table}[t]
  \hspace{-8pt}
  \begin{minipage}[b]{0.6\linewidth}
  	\caption{Comparison with other RoI pooling approaches.}
  	\label{tab:roi_ab}
  	\small
  	\centering
   	\begin{tabular}{cccc}
    	\toprule
    	RoI Method &mAP@0.25 & mAP@0.5 & memory\\
    	\midrule
    	PointRCNN~\cite{shi2019pointrcnn}   & 73.65        & 57.83    & 8,054MB    \\
    	Part-A$^2$~\cite{shi2020p2}         & 74.01        & 58.89   & 6,540MB   \\
    	Ours-SA~\cite{qi2018pointnnetplus}	      & 73.89        & 58.14   & 11,508MB     \\
    	Ours-SpConv		  & \textbf{74.50}        & \textbf{60.31}    & \textbf{2,468MB}  \\
    	\bottomrule
  	\end{tabular}
  \end{minipage}
  \hspace{0pt}
  \begin{minipage}[b]{0.4\linewidth}
  	\caption{Ablate the depth of RoI module.}
  	\label{tab:depth_ab}
  	\small
  	\centering
  	\begin{tabular}{ccc}
    	\toprule
    	$G_t, t=1...$  & mAP@0.25 & mAP@0.5 \\
    	\midrule
    	$\{1\}$		    & 72.33        & 58.42      \\
    	$\{7 , 1\}$		    & \textbf{74.50}        & 60.31      \\
    	$\{7 , 5 , 1\}$		    & 73.91        & \textbf{60.61}      \\
    	$\{7 , 5 , 3 , 1\}$		    & 73.72        & 59.50      \\
    	\bottomrule
  	\end{tabular}
  \end{minipage}
\end{table}

\begin{table}[t]
  \begin{minipage}[b]{0.5\linewidth}
  	\caption{Effect of different proposal sampling resolution $G_x \times G_y \times G_z$ in the $1^{st}$ sparse abstraction.}
  	\label{tab:partsize_ab}
  	\centering
  	\begin{tabular}{ccc}
    	\toprule
    	$G_x \times G_y \times G_z$ & mAP@0.25 & mAP@0.5\\
    	\midrule
    	$3 \times 3 \times 3$		    & 73.56        & 59.61      \\
    	$5 \times 5 \times 5$		    & 74.21        & 60.09      \\
    	$7 \times 7 \times 7$		    & \textbf{74.50}        & 60.31      \\
    	$9 \times 9 \times 9$		    & 74.41        & 60.43      \\
    	$11 \times 11 \times 11$		    & 74.44        & \textbf{60.49}      \\
	
    	\bottomrule
  	\end{tabular}
  \end{minipage}
  \hspace{20pt}
  \begin{minipage}[b]{0.45\linewidth}
  	\caption{Effect of using different grouping kernel sizes $k^{(p)}$ in the $1^{st}$ sparse abstraction.}
  	\label{tab:pk_ab}
  	\centering
  	\begin{tabular}{ccc}
    	\toprule
    	$k^{(p)}$  & mAP@0.25 & mAP@0.5 \\
    	\midrule
    	$3 \times 3 \times 3$		    & 73.87        & 59.82      \\
    	$5 \times 5 \times 5$		    & 74.50        & 60.31      \\
    	$7 \times 7 \times 7$		    & 74.47        & 60.39      \\
    	$9 \times 9 \times 9$		    & \textbf{74.52}        & 60.42      \\
    	$11 \times 11 \times 11$		& 74.45        & \textbf{60.46}      \\
	
    	\bottomrule
  	\end{tabular}
  \end{minipage}
\end{table}

\section{Conclusion}\label{sec:conclusion}
In this paper, we propose CAGroup3D, a two-stage fully convolutional 3D object detector, which generates some 3D proposals by utilizing the class-aware local grouping module on the object voxels with same semantic predictions. Then, to efficiently recover the features of the missed voxels due to incorrect semantic segmentation, we design a fully sparse convolutional RoI pooling module, which is memory-and-computation efficient and could better encode spatial information than previous max-pooling based RoI methods. Equipped with the above designs, our model achieves state-of-the-art on ScanNet V2 and SUN RGB-D benchmarks with remarkable performance gains.

\textbf{Limitations.} CAGroup3D mainly focuses on the inter-category locality, which is class-specific and diverse
among the different classes, but ignores the intra-category discriminations. Due to the incompleteness of the point cloud and the scale variance within the classes, the object spatial dimension of the same class is also variational, which leads to the diverse intra-category locality. Although our grouping algorithm can implicitly handle this problem in some degree by the learnable convlutional aggregation module, it is still an open problem and will be studied in the future.

\section*{Acknowledgement}
Liwei Wang is supported by National Science Foundation of China (NSFC62276005), The Major Key Project of PCL (PCL2021A12), Exploratory Research Project of Zhejiang Lab (No. 2022RC0AN02), and Project 2020BD006 supported by PKUBaidu Fund. We gratefully acknowledge the support of MindSpore\footnote{https://www.mindspore.cn/}.
\bibliographystyle{plainnat}
\bibliography{reference}

\begin{thebibliography}{54}
\providecommand{\natexlab}[1]{#1}
\providecommand{\url}[1]{\texttt{#1}}
\expandafter\ifx\csname urlstyle\endcsname\relax
  \providecommand{\doi}[1]{doi: #1}\else
  \providecommand{\doi}{doi: \begingroup \urlstyle{rm}\Url}\fi

\bibitem[Azuma(1997)]{azuma1997survey}
Ronald~T Azuma.
\newblock A survey of augmented reality.
\newblock \emph{Presence: teleoperators \& virtual environments}, 1997.

\bibitem[Bansal et~al.(2018)Bansal, Krizhevsky, and
  Ogale]{bansal2018chauffeurnet}
Mayank Bansal, Alex Krizhevsky, and Abhijit Ogale.
\newblock Chauffeurnet: Learning to drive by imitating the best and
  synthesizing the worst.
\newblock 2018.

\bibitem[Billinghurst et~al.(2015)Billinghurst, Clark, and
  Lee]{billinghurst2015survey}
Mark Billinghurst, Adrian Clark, and Gun Lee.
\newblock A survey of augmented reality.
\newblock 2015.

\bibitem[Chen et~al.(2020)Chen, Lei, Song, Ying, Chen, and
  Wu]{chen2020hierarchical}
Jintai Chen, Biwen Lei, Qingyu Song, Haochao Ying, Danny~Z Chen, and Jian Wu.
\newblock A hierarchical graph network for 3d object detection on point clouds.
\newblock In \emph{CVPR}, 2020.

\bibitem[Cheng et~al.(2021)Cheng, Sheng, Shi, Yang, and Xu]{cheng2021back}
Bowen Cheng, Lu~Sheng, Shaoshuai Shi, Ming Yang, and Dong Xu.
\newblock Back-tracing representative points for voting-based 3d object
  detection in point clouds.
\newblock In \emph{CVPR}, 2021.

\bibitem[Choy et~al.(2019)Choy, Gwak, and Savarese]{choy20194d}
Christopher Choy, JunYoung Gwak, and Silvio Savarese.
\newblock 4d spatio-temporal convnets: Minkowski convolutional neural networks.
\newblock In \emph{CVPR}, 2019.

\bibitem[Dai et~al.(2017)Dai, Chang, Savva, Halber, Funkhouser, and
  Nie{\ss}ner]{dai2017scannet}
Angela Dai, Angel~X. Chang, Manolis Savva, Maciej Halber, Thomas Funkhouser,
  and Matthias Nie{\ss}ner.
\newblock Scannet: Richly-annotated 3d reconstructions of indoor scenes.
\newblock In \emph{CVPR}, 2017.

\bibitem[Danila~Rukhovich(2021)]{rukhovich2021fcaf3d}
Anton~Konushin Danila~Rukhovich, Anna~Vorontsova.
\newblock Fcaf3d: Fully convolutional anchor-free 3d object detection.
\newblock \emph{arXiv preprint arXiv:2112.00322}, 2021.

\bibitem[Deng et~al.(2021)Deng, Shi, Li, Zhou, Zhang, and Li]{deng2020voxel}
Jiajun Deng, Shaoshuai Shi, Peiwei Li, Wengang Zhou, Yanyong Zhang, and
  Houqiang Li.
\newblock Voxel r-cnn: Towards high performance voxel-based 3d object
  detection.
\newblock In \emph{AAAI}, 2021.

\bibitem[Engelmann et~al.(2020)Engelmann, Bokeloh, Fathi, Leibe, and
  Nie{\ss}ner]{engelmann20203d}
Francis Engelmann, Martin Bokeloh, Alireza Fathi, Bastian Leibe, and Matthias
  Nie{\ss}ner.
\newblock 3d-mpa: Multi-proposal aggregation for 3d semantic instance
  segmentation.
\newblock In \emph{CVPR}, 2020.

\bibitem[Fan et~al.(2022)Fan, Pang, Zhang, Wang, Zhao, Wang, Wang, and
  Zhang]{fan2021embracing}
Lue Fan, Ziqi Pang, Tianyuan Zhang, Yu-Xiong Wang, Hang Zhao, Feng Wang, Naiyan
  Wang, and Zhaoxiang Zhang.
\newblock Embracing single stride 3d object detector with sparse transformer.
\newblock In \emph{CVPR}, 2022.

\bibitem[Graham and van~der Maaten(2017)]{graham2017submanifold}
Benjamin Graham and Laurens van~der Maaten.
\newblock Submanifold sparse convolutional networks.
\newblock \emph{arXiv preprint arXiv:1706.01307}, 2017.

\bibitem[Graham et~al.(2018)Graham, Engelcke, and Van Der~Maaten]{graham20183d}
Benjamin Graham, Martin Engelcke, and Laurens Van Der~Maaten.
\newblock 3d semantic segmentation with submanifold sparse convolutional
  networks.
\newblock In \emph{CVPR}, 2018.

\bibitem[Gwak et~al.(2020)Gwak, Choy, and Savarese]{gwak2020generative}
JunYoung Gwak, Christopher Choy, and Silvio Savarese.
\newblock Generative sparse detection networks for 3d single-shot object
  detection.
\newblock In \emph{ECCV}, 2020.

\bibitem[He et~al.(2016)He, Zhang, Ren, and Sun]{he2016deep}
Kaiming He, Xiangyu Zhang, Shaoqing Ren, and Jian Sun.
\newblock Deep residual learning for image recognition.
\newblock In \emph{CVPR}, 2016.

\bibitem[Hong et~al.(2021)Hong, Pan, Sun, and Jia]{hong2021deep}
Yuanduo Hong, Huihui Pan, Weichao Sun, and Yisong Jia.
\newblock Deep dual-resolution networks for real-time and accurate semantic
  segmentation of road scenes.
\newblock \emph{arXiv preprint arXiv:2101.06085}, 2021.

\bibitem[Hou et~al.(2019)Hou, Dai, and Nie{\ss}ner]{hou20193d}
Ji~Hou, Angela Dai, and Matthias Nie{\ss}ner.
\newblock 3d-sis: 3d semantic instance segmentation of rgb-d scans.
\newblock In \emph{CVPR}, 2019.

\bibitem[Jin et~al.(2021)Jin, Zhang, Wang, and Wang]{jin2021nonconvex}
Jikai Jin, Bohang Zhang, Haiyang Wang, and Liwei Wang.
\newblock Non-convex distributionally robust optimization: Non-asymptotic
  analysis.
\newblock In \emph{NeurIPS}, 2021.

\bibitem[Lang et~al.(2019)Lang, Vora, Caesar, Zhou, Yang, and
  Beijbom]{lang2019pointpillars}
Alex~H Lang, Sourabh Vora, Holger Caesar, Lubing Zhou, Jiong Yang, and Oscar
  Beijbom.
\newblock Pointpillars: Fast encoders for object detection from point clouds.
\newblock In \emph{CVPR}, 2019.

\bibitem[Lin et~al.(2017{\natexlab{a}})Lin, Doll{\'a}r, Girshick, He,
  Hariharan, and Belongie]{lin2017feature}
Tsung-Yi Lin, Piotr Doll{\'a}r, Ross Girshick, Kaiming He, Bharath Hariharan,
  and Serge Belongie.
\newblock Feature pyramid networks for object detection.
\newblock In \emph{CVPR}, 2017{\natexlab{a}}.

\bibitem[Lin et~al.(2017{\natexlab{b}})Lin, Goyal, Girshick, He, and
  Doll{\'a}r]{lin2017focal}
Tsung-Yi Lin, Priya Goyal, Ross Girshick, Kaiming He, and Piotr Doll{\'a}r.
\newblock Focal loss for dense object detection.
\newblock In \emph{ICCV}, 2017{\natexlab{b}}.

\bibitem[Liu et~al.(2021)Liu, Zhang, Cao, Hu, and Tong]{liu2021group}
Ze~Liu, Zheng Zhang, Yue Cao, Han Hu, and Xin Tong.
\newblock Group-free 3d object detection via transformers.
\newblock In \emph{ICCV}, 2021.

\bibitem[Loshchilov and Hutter(2019)]{loshchilov2018decoupled}
Ilya Loshchilov and Frank Hutter.
\newblock Decoupled weight decay regularization.
\newblock In \emph{ICLR}, 2019.

\bibitem[Misra et~al.(2021)Misra, Girdhar, and Joulin]{misra2021end}
Ishan Misra, Rohit Girdhar, and Armand Joulin.
\newblock An end-to-end transformer model for 3d object detection.
\newblock In \emph{CVPR}, 2021.

\bibitem[Qi et~al.(2017{\natexlab{a}})Qi, Su, Mo, and Guibas]{qi2017pointnet}
Charles~R Qi, Hao Su, Kaichun Mo, and Leonidas~J Guibas.
\newblock Pointnet: Deep learning on point sets for 3d classification and
  segmentation.
\newblock In \emph{CVPR}, 2017{\natexlab{a}}.

\bibitem[Qi et~al.(2018)Qi, Liu, Wu, Su, and Guibas]{qi2018frustum}
Charles~R Qi, Wei Liu, Chenxia Wu, Hao Su, and Leonidas~J Guibas.
\newblock Frustum pointnets for 3d object detection from rgb-d data.
\newblock In \emph{CVPR}, 2018.

\bibitem[Qi et~al.(2019)Qi, Litany, He, and Guibas]{qi2019deep}
Charles~R Qi, Or~Litany, Kaiming He, and Leonidas~J Guibas.
\newblock Deep hough voting for 3d object detection in point clouds.
\newblock In \emph{ICCV}, 2019.

\bibitem[Qi et~al.(2020)Qi, Chen, Litany, and Guibas]{qi2020imvotenet}
Charles~R Qi, Xinlei Chen, Or~Litany, and Leonidas~J Guibas.
\newblock Imvotenet: Boosting 3d object detection in point clouds with image
  votes.
\newblock In \emph{CVPR}, 2020.

\bibitem[Qi et~al.(2017{\natexlab{b}})Qi, Yi, Su, and
  Guibas]{qi2018pointnnetplus}
Charles~Ruizhongtai Qi, Li~Yi, Hao Su, and Leonidas~J Guibas.
\newblock Pointnet++: Deep hierarchical feature learning on point sets in a
  metric space.
\newblock In \emph{NeurIPS}, 2017{\natexlab{b}}.

\bibitem[Shi et~al.(2019)Shi, Wang, and Li]{shi2019pointrcnn}
Shaoshuai Shi, Xiaogang Wang, and Hongsheng Li.
\newblock Pointrcnn: 3d object proposal generation and detection from point
  cloud.
\newblock In \emph{CVPR}, 2019.

\bibitem[Shi et~al.(2020{\natexlab{a}})Shi, Guo, Jiang, Wang, Shi, Wang, and
  Li]{shi2020pv}
Shaoshuai Shi, Chaoxu Guo, Li~Jiang, Zhe Wang, Jianping Shi, Xiaogang Wang, and
  Hongsheng Li.
\newblock Pv-rcnn: Point-voxel feature set abstraction for 3d object detection.
\newblock In \emph{CVPR}, 2020{\natexlab{a}}.

\bibitem[Shi et~al.(2020{\natexlab{b}})Shi, Wang, Shi, Wang, and Li]{shi2020p2}
Shaoshuai Shi, Zhe Wang, Jianping Shi, Xiaogang Wang, and Hongsheng Li.
\newblock From points to parts: 3d object detection from point cloud with
  part-aware and part-aggregation network.
\newblock \emph{TPAMI}, 2020{\natexlab{b}}.

\bibitem[Song et~al.(2015)Song, Lichtenberg, and Xiao]{sunrgbd}
Shuran Song, Samuel~P. Lichtenberg, and Jianxiong Xiao.
\newblock Sun rgb-d: A rgb-d scene understanding benchmark suite.
\newblock In \emph{CVPR}, 2015.

\bibitem[Sun et~al.(2019)Sun, Xiao, Liu, and Wang]{sun2019deep}
Ke~Sun, Bin Xiao, Dong Liu, and Jingdong Wang.
\newblock Deep high-resolution representation learning for human pose
  estimation.
\newblock In \emph{CVPR}, 2019.

\bibitem[Vu et~al.(2022)Vu, Kim, Luu, Nguyen, and Yoo]{vu2022softgroup}
Thang Vu, Kookhoi Kim, Tung~M Luu, Thanh Nguyen, and Chang~D Yoo.
\newblock Softgroup for 3d instance segmentation on point clouds.
\newblock In \emph{CVPR}, 2022.

\bibitem[Wang et~al.(2019)Wang, Devin, Cai, Kr{\"a}henb{\"u}hl, and
  Darrell]{wang2019monocular}
Dequan Wang, Coline Devin, Qi-Zhi Cai, Philipp Kr{\"a}henb{\"u}hl, and Trevor
  Darrell.
\newblock Monocular plan view networks for autonomous driving.
\newblock In \emph{IROS}, 2019.

\bibitem[Wang et~al.(2021)Wang, Wang, Zhu, Dai, and
  Wang]{wang2021collaborative}
Haiyang Wang, Wenguan Wang, Xizhou Zhu, Jifeng Dai, and Liwei Wang.
\newblock Collaborative visual navigation.
\newblock \emph{arXiv preprint arXiv:2107.01151}, 2021.

\bibitem[Wang et~al.(2022{\natexlab{a}})Wang, Shi, Yang, Fang, Qian, Li,
  Schiele, and Wang]{wang2022rbgnet}
Haiyang Wang, Shaoshuai Shi, Ze~Yang, Rongyao Fang, Qi~Qian, Hongsheng Li,
  Bernt Schiele, and Liwei Wang.
\newblock Rbgnet: Ray-based grouping for 3d object detection.
\newblock In \emph{CVPR}, 2022{\natexlab{a}}.

\bibitem[Wang et~al.(2020)Wang, Sun, Cheng, Jiang, Deng, Zhao, Liu, Mu, Tan,
  Wang, et~al.]{wang2020deep}
Jingdong Wang, Ke~Sun, Tianheng Cheng, Borui Jiang, Chaorui Deng, Yang Zhao,
  Dong Liu, Yadong Mu, Mingkui Tan, Xinggang Wang, et~al.
\newblock Deep high-resolution representation learning for visual recognition.
\newblock \emph{TPAMI}, 2020.

\bibitem[Wang et~al.(2022{\natexlab{b}})Wang, Chen, Cao, Huang, Sun, and
  Wang]{wang2022multimodal}
Yikai Wang, Xinghao Chen, Lele Cao, Wenbing Huang, Fuchun Sun, and Yunhe Wang.
\newblock Multimodal token fusion for vision transformers.
\newblock In \emph{CVPR}, 2022{\natexlab{b}}.

\bibitem[Xie et~al.(2020)Xie, Lai, Wu, Wang, Zhang, Xu, and
  Wang]{xie2020mlcvnet}
Qian Xie, Yu-Kun Lai, Jing Wu, Zhoutao Wang, Yiming Zhang, Kai Xu, and Jun
  Wang.
\newblock Mlcvnet: Multi-level context votenet for 3d object detection.
\newblock In \emph{CVPR}, 2020.

\bibitem[Xie et~al.(2021)Xie, Lai, Wu, Wang, Lu, Wei, and Wang]{xie2021venet}
Qian Xie, Yu-Kun Lai, Jing Wu, Zhoutao Wang, Dening Lu, Mingqiang Wei, and Jun
  Wang.
\newblock Venet: Voting enhancement network for 3d object detection.
\newblock In \emph{ICCV}, 2021.

\bibitem[Xu et~al.(2022)Xu, Dong, Ding, Wang, Xu, and Li]{xu2022fusionrcnn}
Xinli Xu, Shaocong Dong, Lihe Ding, Jie Wang, Tingfa Xu, and Jianan Li.
\newblock Fusionrcnn: Lidar-camera fusion for two-stage 3d object detection.
\newblock \emph{arXiv preprint arXiv:2209.10733}, 2022.

\bibitem[Yan et~al.(2018)Yan, Mao, and Li]{yan2018second}
Yan Yan, Yuxing Mao, and Bo~Li.
\newblock Second: Sparsely embedded convolutional detection.
\newblock \emph{Sensors}, 2018.

\bibitem[Yang et~al.(2018)Yang, Luo, and Urtasun]{yang2018pixor}
Bin Yang, Wenjie Luo, and Raquel Urtasun.
\newblock Pixor: Real-time 3d object detection from point clouds.
\newblock In \emph{CVPR}, 2018.

\bibitem[Yang et~al.(2022)Yang, Shi, Chen, and Wang]{yang2022boosting}
Hao Yang, Chen Shi, Yihong Chen, and Liwei Wang.
\newblock Boosting 3d object detection via object-focused image fusion.
\newblock \emph{arXiv preprint arXiv:2207.10589}, 2022.

\bibitem[Yang et~al.(2019)Yang, Sun, Liu, Shen, and Jia]{yang2019std}
Zetong Yang, Yanan Sun, Shu Liu, Xiaoyong Shen, and Jiaya Jia.
\newblock Std: Sparse-to-dense 3d object detector for point cloud.
\newblock In \emph{ICCV}, 2019.

\bibitem[Yi et~al.(2019)Yi, Zhao, Wang, Sung, and Guibas]{yi2018gspn}
Li~Yi, Wang Zhao, He~Wang, Minhyuk Sung, and Leonidas Guibas.
\newblock Gspn: Generative shape proposal network for 3d instance segmentation
  in point cloud.
\newblock In \emph{CVPR}, 2019.

\bibitem[Yin et~al.(2021)Yin, Zhou, and Krahenbuhl]{yin2021cvpr}
Tianwei Yin, Xingyi Zhou, and Philipp Krahenbuhl.
\newblock Center-based 3d object detection and tracking.
\newblock In \emph{CVPR}, 2021.

\bibitem[Zhang et~al.(2020{\natexlab{a}})Zhang, Jin, Fang, and
  Wang]{zhang2020_b282d173}
Bohang Zhang, Jikai Jin, Cong Fang, and Liwei Wang.
\newblock Improved analysis of clipping algorithms for non-convex optimization.
\newblock In \emph{NeurIPS}, 2020{\natexlab{a}}.

\bibitem[Zhang et~al.(2020{\natexlab{b}})Zhang, Sun, Yang, and
  Huang]{zhang2020h3dnet}
Zaiwei Zhang, Bo~Sun, Haitao Yang, and Qixing Huang.
\newblock H3dnet: 3d object detection using hybrid geometric primitives.
\newblock In \emph{ECCV}, 2020{\natexlab{b}}.

\bibitem[Zheng et~al.(2022)Zheng, Duan, Lu, Zhou, and
  Tian]{zheng2022hyperdet3d}
Yu~Zheng, Yueqi Duan, Jiwen Lu, Jie Zhou, and Qi~Tian.
\newblock Hyperdet3d: Learning a scene-conditioned 3d object detector.
\newblock In \emph{CVPR}, 2022.

\bibitem[Zhou and Tuzel(2018)]{zhou2018voxelnet}
Yin Zhou and Oncel Tuzel.
\newblock Voxelnet: End-to-end learning for point cloud based 3d object
  detection.
\newblock In \emph{CVPR}, 2018.

\bibitem[Zhu et~al.(2017)Zhu, Mottaghi, Kolve, Lim, Gupta, Fei-Fei, and
  Farhadi]{zhu2017target}
Yuke Zhu, Roozbeh Mottaghi, Eric Kolve, Joseph~J Lim, Abhinav Gupta,
  Li~Fei-Fei, and Ali Farhadi.
\newblock Target-driven visual navigation in indoor scenes using deep
  reinforcement learning.
\newblock In \emph{ICRA}, 2017.

\end{thebibliography}

\section*{Checklist}

\begin{enumerate}

\item For all authors...
\begin{enumerate}
  \item Do the main claims made in the abstract and introduction accurately reflect the paper's contributions and scope?
    \answerYes{}
  \item Did you describe the limitations of your work?
    \answerYes{See Section \ref{sec:conclusion}.}
  \item Did you discuss any potential negative societal impacts of your work?
    \answerNo{Our work is only for academic research purpose.}
  \item Have you read the ethics review guidelines and ensured that your paper conforms to them?
    \answerYes{}
\end{enumerate}

\item If you are including theoretical results...
\begin{enumerate}
  \item Did you state the full set of assumptions of all theoretical results?
    \answerNA{}
        \item Did you include complete proofs of all theoretical results?
    \answerNA{}
\end{enumerate}

\item If you ran experiments...
\begin{enumerate}
  \item Did you include the code, data, and instructions needed to reproduce the main experimental results (either in the supplemental material or as a URL)?
    \answerYes{Our implementation details are posted on Section \ref{sec:imple} and the supplemental materials. The data are public datasets and our code will be available.}
  \item Did you specify all the training details (e.g., data splits, hyperparameters, how they were chosen)?
    \answerYes{See Section ~\ref{sec:imple} and Appendix.}
        \item Did you report error bars (e.g., with respect to the random seed after running experiments multiple times)?
    \answerYes{We report both the best and average performance of 25 trials by default.}
        \item Did you include the total amount of compute and the type of resources used (e.g., type of GPUs, internal cluster, or cloud provider)?
    \answerYes{ See Section \ref{sec:imple}.}
\end{enumerate}

\item If you are using existing assets (e.g., code, data, models) or curating/releasing new assets...
\begin{enumerate}
  \item If your work uses existing assets, did you cite the creators?
    \answerYes{ScanNet V2~\cite{dai2017scannet} and SUN RGB-D~\cite{sunrgbd}}
  \item Did you mention the license of the assets?
    \answerNA{}
  \item Did you include any new assets either in the supplemental material or as a URL?
    \answerNo{We don't include any new assets.}
  \item Did you discuss whether and how consent was obtained from people whose data you're using/curating?
    \answerYes{ScanNet V2~\cite{dai2017scannet} and SUN RGB-D~\cite{sunrgbd} are two standard benchmarks.}
  \item Did you discuss whether the data you are using/curating contains personally identifiable information or offensive content?
    \answerYes{ScanNet V2~\cite{dai2017scannet} and SUN RGB-D~\cite{sunrgbd} are two standard benchmarks.}
\end{enumerate}

\item If you used crowdsourcing or conducted research with human subjects...
\begin{enumerate}
  \item Did you include the full text of instructions given to participants and screenshots, if applicable?
    \answerNA{We didn't use crowdsourcing or conduct research with human subjects...}
  \item Did you describe any potential participant risks, with links to Institutional Review Board (IRB) approvals, if applicable?
    \answerNA{We didn't use crowdsourcing or conduct research with human subjects...}
  \item Did you include the estimated hourly wage paid to participants and the total amount spent on participant compensation?
    \answerNA{We didn't use crowdsourcing or conduct research with human subjects...}
\end{enumerate}

\end{enumerate}

\appendix
In the supplementary material, we first provide more implementation details of the network architecture (\S \ref{sec:detail}), then present the per-category evaluation (\S \ref{sec:per-class}), latency and runtime memory analysis (\S \ref{sec:runtime}), more ablation studies (\S \ref{sec:more_ab}) and visualization of quantitative results (\S \ref{sec:vis}).

\section{Implementation Details}
\label{sec:detail}
As mentioned in the main paper, the CAGroup3D architecture consists of a backbone with dual resolution named BiResNet, a class-aware 3D proposal generation module and a RoI-Conv refinement module. We first detail the backbone and proposal generation module, then elaborate the RoI-Conv refinement module as well as its competitors, and finally present the details of our loss functions.

\subsection{BiResNet Backbone}
Our backbone network is built upon MinkowskiEngine~\cite{choy20194d}, an auto-differentiation library for sparse tensors. In all experiments, we voxelize the original point clouds into sparse tensors with a voxel size of 0.02m and feed them into the backbone network. BiResNet contains two branches, one is the sparse modification of ResNet18~\cite{he2016deep} to extract pyramid contextual features with proper downsampling modules, the other one is an auxiliary branch to hold a high-resolution feature map whose resolution is 1/2 of the input 3D voxels. To achieve information interaction between the two streams, we construct a bilateral fusion block, which includes fusing the high-resolution branch into the low-resolution (high-to-low) and low-resolution into high-resolution (low-to-high). As for high-to-low fusion, high-resolution features are downsampled by a sparse convolution block with a specifical stride (\textit{e.g.}, 2 and 4 for different stages) before being added to the low-resolution feature map. Meanwhile, an interpolation operation and another channel-compression convolution are used to upsample the low-resolution feature map before being fused with the high-resolution auxiliary branch. All convolution layers are followed by batch or instance normalization and ReLU activation function. The output of the backbone network are 64-dimensional voxel-wise latent features.

\subsection{Class-aware 3D Proposal Generation Module}
The class-aware 3D proposal generation module consists of a semantic and vote prediction module, a class-aware local grouping module and an anchor-free proposal head. The detailed computation procedure is provided in Algorithm \ref{algo:calg}. The proposal head comprises three parallel sparse convolutional layers with weights shared across all class-individual feature maps. For each candidate object, theses layers output classification probabilities for each class, bounding box parameters and 3D centerness values separately. Finally, we filter out those proposal bounding boxes with score less than 0.01, then apply oriented NMS with 3D IoU threshold of 0.5 to remove overlapped bounding boxes and reduce the number of proposals.  

\begin{algorithm} 
  \caption{Algorithm of Class-Aware 3D proposal Generation Module.}
  \begin{algorithmic}[1] \label{algo:calg}
      \Statex \textbf{Input:} ~Seed voxels $\{o_i\}_{i=1}^N$, semantic threshold $\tau$, kernel sizes of aggregation $k^{(a)}$,
      \Statex $\quad\quad$ ~~~~Voxel sizes of different classes $\{d_j\}^{N_{class}}_{j=1}$, scale factor $\alpha$.
      \Statex \textbf{Output:} Proposals $P$
      \State Initialize class-aware aggregation results $A = \{\}$.
      
      \Statex {\color{blue}/*Voxel-wise Semantic and Vote Prediction*/}
      \State $\{p_i\}_{i=1}^N = \{\text{MLP}^{\text{vote}}(o_i)\}_{i=1}^N$, ~~~~~~~$\{s_i\}_{i=1}^N = \{\text{MLP}^{\text{sem}}(o_i) \in [0,1]^{N_{class}}\}_{i=1}^N$ 
      \Statex {\color{blue}/*Class-Aware Local Grouping*/}
      \For{$j\leftarrow 0$ to $N_{class}$}
          \Statex ~~~~~~~{\color{blue}/*Slice a semantic subset with $\tau$*/}
          \State $c_j=\{p_i: s_i^j > \tau , i=1, ..., N \}$
          \Statex ~~~~~~~{\color{blue}/*Class-Aware Re-voxelization*/}
          \vspace{4pt}
          \State $\{v_i\}_{i=1}^{\mid V_j \mid} =\text{VFE}(c_j, ~\alpha \cdot d_j, ~Avg)$
          \Statex ~~~~~~~{\color{blue}/*Class-dependent SpConv Aggregation with $k^{(a)}$*/}
          \State $A^{(j)} = \{a_i^{(j)} \mid a_i^{(j)} = \text{SparseConv}_{\text{3D}}^{(j)}(v_i, ~ \{v_i\}_{i=1}^{\mid V_j \mid}, ~k^{(a)})\}_{i=1}^{\mid V_j \mid}$
          \Statex ~~~~~~~{\color{blue}/*Merge the subset*/}
          \State  $A$ append $A^{(j)}$  \Comment{Not a unique OP that each loc may have multiple features.}
      \EndFor 
      
      \Statex {\color{blue}/*Proposal Head with NMS*/}
      \State $P = \text{NMS}\left(\{\text{MLP}(A_l)\}_{l=0}^{\mid V_0 \mid + ... +\mid V_{N_{class}} \mid}\right)$
      \State \textbf{Return $P$}
  \end{algorithmic}
\end{algorithm} 

\subsection{RoI-Conv Refinement Module}
Given the proposals of Stage-I, we further select 128 proposals whose 3D IoU with ground truth are greater than 0.3 as training samples for each scene, while reserve all proposals during inference. Finally, the proposals and voxel features from backbone are fed into RoI-Conv pooling module with two stacked sparse abstraction blocks to obtain the RoI-specific features. We also provide more implementation details of other RoI pooling strategies mentioned in the main paper.

\textbf{PointRCNN}. We first slightly enlarge the proposals by 0.3m, then randomly take out 128 voxels $\{l_n\}_{n=1}^{128}$ within each proposal for further processing. These cropped voxels are regarded as input points and fed into the hierarchical PointNet++ ~\cite{qi2018pointnnetplus} with two SA layers to obtain the final RoI-specific features. The first SA layer uses farthest point sampling (FPS) to sample 32 key points from the input and applies a set abstraction operation~\cite{qi2017pointnet} centered on each key points to encode local patterns. The radius and number of neighbors are set to 0.4m and 16. Finally the sampled key points are pooled to a feature vector by the last SA layer for further proposal refinement. 

\textbf{Part-A$^2$}. Instead of directly processing the irregular points within proposals as PointRCNN, Part-A$^2$ converts the contiguously distributed points into regular voxels with a fixed spatial shape, where the average pooling operation is adopted to pool the points in the same voxel. For a fair comparison, we adopt the same spatial shape (\textit{i.e.}, $7 \times 7 \times 7$) as ours, and then several sparse convolutions are stacked to aggregate all part features into a feature vector. Notably, we follow the original paper~\cite{shi2020p2} and keep the empty voxels in each proposal to encode the bounding box's geometric information.

\textbf{Ours-SA}. In this variant, we replace the sparse convolution operation used for encoding local patterns in our sparse abstraction block with set abstraction~\cite{qi2018pointnnetplus}. Specifically, given the RoI-specific points set $\widetilde{\mathcal{G}}=\{g_k\}^{\mid \widetilde{\mathcal{G}} \mid}_{k=1}$ sampled from the proposals, instead of exploiting sparse convolution centered on each points, we adopt ball query to cover neighboring voxels. Then a PointNet operation is applied on each query group to learn the local patterns. We follow the same two-layers architecture and proposal sampling resolutions as our RoI-Conv module.  Their corresponding radius and number of neighbors are set to (0.3m, 2.0m) and (16, $7 \times 7 \times 7$) respectively.  

As mentioned in the main paper, we compare our RoI-Conv module with the above three variants both on detection scores and computation cost. Note that the computation cost is measured by training memory with the batch size of 8. The experiments show that our RoI-Conv module has significant superiority. All the experiments are run on the same workstation and environment.

We further explain how we change the depth of RoI module introduced in the main paper. To be specific, we stack different number of sparse abstraction blocks with fixed sparse kernel size $k^{(p)} = 5$ and decreasing proposal sampling resolutions $\{G_t\}, t=1,...,n$, where $n$ is the number of blocks. For example, $\{G_t\}=\{1\}$ means we only sample one grid point (proposal center) for each proposal and aggregate input voxels from backbone by directly applying sparse convolution centered on these points; $\{G_t\}=\{7, 5, 1\}$ means the first sparse abstraction block outputs a voxel set $\mathcal{Q}^1$ where the voxels are sampled from the proposals with $7 \times 7 \times 7$ resolution and serve as convolution centers to encode their local patterns from the input voxels. 
The second sparse abstraction block further samples a smaller voxel set $\mathcal{Q}^2$ from the proposals with $5 \times 5 \times 5$ resolution, and similarly obtains the voxelwise output features by applying sparse convolution centered on $\mathcal{Q}^2$ to cover their neighbors from $\mathcal{Q}^1$. Finally, $\mathcal{Q}^2$ is fed into the last sparse abstraction block. We use a sparse convolution with $k^{(p)}=5$ to aggregate all the part features into the proposal center and get the RoI-specific feature vector. Other settings can be easily understood by analogy with the above explanation. Note that the sparse kernel size $k^{(p)}$ in the last block is equal to the proposal sampling resolution in the second-to-last block to aggregate all information in the proposal. 

\subsection{Loss Function Details}
\label{sec:loss}
Our model is trained end-to-end with a multi-task loss including semantic loss $\mathcal{L}_\text{sem}$, voting loss $\mathcal{L}_{\text{vote-reg}}$, centerness loss $\mathcal{L}_\text{cntr}$, bounding box estimation loss $\mathcal{L}_\text{box}$, classification losses $\mathcal{L}_\text{cls}$ for Stage-I and bbox refinement loss $\mathcal{L}_\text{rebox}$ for Stage-II.

\begin{equation}
    \begin{aligned}
    L = \beta_\text{sem}\mathcal{L}_\text{sem} + \beta_\text{vote}\mathcal{L}_{\text{vote}} + \beta_\text{cntr}\mathcal{L}_\text{cntr} \\ 
    +\beta_\text{box}\mathcal{L}_\text{box} + \beta_\text{cls}\mathcal{L}_\text{cls} + \beta_\text{rebox}\mathcal{L}_\text{rebox}.
    \end{aligned}
\end{equation}
The second stage loss $\mathcal{L}_\text{rebox}$ consists of a regression loss $\mathcal{L}_{\text{smooth-}\ell_1}$ and a iou loss $\mathcal{L}_{\text{iou}}$. For the regression loss, both the 3D proposals and their corresponding ground-truth bounding boxes are transformed into the canonical coordinate systems, which means the 3D proposal $b_i = (x_i, y_i, z_i, h_i, w_i, l_i, \theta_i)$ and ground-truth bounding box $b_i^{gt}= (x_i^{gt}, y_i^{gt}, z_i^{gt}, h_i^{gt}, w_i^{gt}, l_i^{gt}, \theta_i^{gt})$ would be transformed to 
\begin{equation}
\begin{aligned}
    & ~~~~~~~~~~~~~~~~~~~~~~~~~\tilde{b}_i = (0,~ 0,~ 0, ~h_i, ~w_i, ~l_i, ~0), \\
    & \tilde{b}_i^{gt} = (x_i^{gt} - x_i,~ y_i^{gt} - y_i,~ z_i^{gt} - z_i,~ h_i^{gt},~ w_i^{gt},~ l_i^{gt}, ~\theta_i^{gt} - \theta_i).
\end{aligned}
\end{equation}
Then following the traditional residual learning method and sin-cos heading encoding strategy, we obtain the final target $t$ as follow:
\begin{equation}
    t = (\frac{x_i^{gt} - x_i}{d},~ \frac{y_i^{gt} - y_i}{d},~ \frac{z_i^{gt} - z_i}{d},~ \text{log}(\frac{h_i^{gt}}{h_i}),~ \text{log}(\frac{w_i^{gt}}{w_i}),~ \text{log}(\frac{l_i^{gt}}{l_i}), ~\text{sin}(\Delta_{\theta}), ~\text{cos}(\Delta_{\theta})),
\end{equation}
where $d = \sqrt{h_i^2 + w_i^2 + l_i^2},~ \Delta_{\theta} = \theta_i^{gt} - \theta_i$. Finally the smooth-$\mathcal{L}_1$ loss is adopted to compute the regression loss. For the iou loss, we get the final refined bounding boxes decoded from the prediction logits and compute their rotated IoU with ground-truth bouding boxes as used in Stage-I. 

The balancing factors are set default as $\beta_{sem} = 1.0, \beta_{vote} = 1.0, \beta_{cntr} = 1.0, ~\beta_{box} = 1.0, ~\beta_{cls} = 1.0, ~\beta_{rebox} = 0.5$.

\begin{table*}[h]
\large
\setlength{\abovecaptionskip}{0pt}  
\setlength{\belowcaptionskip}{5pt}
\caption{3D detection scores per category on the ScanNetV2, evaluated with mAP@0.25 IoU.}
\label{tab:25scan}
\renewcommand{\arraystretch}{1.5}{
\resizebox{\textwidth}{!}{
\begin{tabular}{c|c|c|c|c|c|c|c|c|c|c|c|c|c|c|c|c|c|c|c}
    \toprule[1.5pt]
     & cab & bed & chair & sofa & tabl & door & wind & bkshf & pic & cntr & desk & curt & frig & showr & toil & sink & bath & ofurn & mAP \\
    \hline
    VoteNet~\cite{qi2020imvotenet} & 47.87 & 90.79 &90.07 &90.78 &60.22 &53.83 &43.71 &55.56 &12.38 &66.85 &66.02 &52.37 &52.05 &63.94 &97.40 &52.32 &92.57 &43.37 &62.90 \\
    MLCVNet~\cite{xie2020mlcvnet} & 42.50 & 88.50 & 90.00 & 87.40 & 63.50 & 56.90 & 47.00 & 57.00 & 12.00 & 63.90 & 76.10 & 56.70 & 60.90 & 65.90 & 98.30 & 59.20 & 87.20 & 47.90 & 64.50\\ 
    BRNet~\cite{cheng2021back} & 49.90 &88.30 &91.90 &86.90 &69.30 &59.20 &45.90 &52.10 &15.30 &72.00 &76.80 &57.10 &60.40 &73.60 &93.80 &58.80 &92.20 &47.10 &66.10 \\
    H3DNet~\cite{zhang2020h3dnet} & 49.40 & 88.60 & 91.80 & 90.20 & 64.90 & 61.00 & 51.90 & 54.90 & 18.60 & 62.00 & 75.90 & 57.30 & 57.20 & 75.30 & 97.90 & 67.40 & 92.50 & 53.60 & 67.20\\
    Group-free~\cite{liu2021group} &52.10 &92.90 &93.60 &88.00 &\textbf{70.70} &60.70 &53.70 &62.40 &16.10 &58.50 &80.90 &67.90 &47.00 &76.30 &99.60 &72.00 &\textbf{95.30} &56.40 &69.10 \\
    FCAF3D~\cite{rukhovich2021fcaf3d} & 57.20 &87.00 &95.00 &92.30 &70.30 &61.10 &60.20 &64.50 &29.90 &64.30 &71.50 &60.10 &52.40 &\textbf{83.90} &99.90 &\textbf{84.70} &86.60 &65.40 &71.50 \\
    \hline
    Ours & \textbf{60.37} &\textbf{93.00} &\textbf{95.25} &\textbf{92.32} &69.95 &\textbf{67.95} &\textbf{63.60} &\textbf{67.29} &\textbf{40.70} &\textbf{77.01} &\textbf{83.87} &\textbf{69.43} &\textbf{65.65} &73.00 &\textbf{99.97} &79.70 &86.98 &\textbf{66.12} &\textbf{75.12}\\
    \hline
\end{tabular}}}
\vspace{-4mm}
\end{table*}

\begin{table*}[h]
\large
\setlength{\abovecaptionskip}{0pt}  
\setlength{\belowcaptionskip}{5pt}
\caption{3D detection scores per category on the ScanNetV2, evaluated with mAP@0.50 IoU.}
\label{tab:50scan}
\renewcommand{\arraystretch}{1.5}{
\resizebox{\textwidth}{!}{
\begin{tabular}{c|c|c|c|c|c|c|c|c|c|c|c|c|c|c|c|c|c|c|c}
    \toprule[1.5pt]
    & cab & bed & chair & sofa & tabl & door & wind & bkshf & pic & cntr & desk & curt & frig & showr & toil & sink & bath & ofurn & mAP \\
    \hline
    VoteNet~\cite{qi2020imvotenet} &8.10 &76.10 &67.20 &68.80 &42.40 &15.30 &6.40 &28.00 &1.30 &9.50 &37.50 &11.60 &27.80 &10.00 &86.50 &16.80 &78.90 &11.70 &33.50 \\
    BRNet~\cite{cheng2021back} & 28.70 &80.60 &81.90 &80.60 &60.80 &35.50 &22.20 &48.00 &7.50 &\textbf{43.70} &54.80 &39.10 &51.80 &35.90 &88.90 &38.70 &84.40 &33.00 &50.90 \\
    H3DNet~\cite{zhang2020h3dnet} & 20.50 & 79.70 & 80.10 & 79.60 & 56.20 & 29.00 & 21.30 & 45.50 & 4.20 & 33.50 & 50.60 & 37.30 & 41.40 & 37.00 & 89.10 & 35.10 & \textbf{90.20} & 35.40 & 48.10\\
    Group-free~\cite{liu2021group} &26.00 &81.30 &82.90 &70.70 &62.20 &41.70 &26.50 &55.80 &7.80 &34.70 &\textbf{67.20} &43.90 &44.30 &44.10 &92.80 &37.40 &89.70 &40.60 &52.80 \\
    FCAF3D~\cite{rukhovich2021fcaf3d} &35.80 &81.50 &89.80 &85.00 &62.00 &44.10 &30.70 &58.40 &17.90 &31.30 &53.40 &44.20 &46.80 &\textbf{64.20} &91.60 &52.60 &84.50 &57.10 &57.30 \\
    \hline
    Ours & \textbf{41.35} &\textbf{82.82} &\textbf{90.82} &\textbf{85.62} &\textbf{64.93} &\textbf{54.33} &\textbf{37.33} &\textbf{64.10} &\textbf{31.38} &41.08 &63.62 &\textbf{44.38} &\textbf{56.95} &49.26 &\textbf{98.19} &\textbf{55.44} &82.40 &\textbf{58.82} &\textbf{61.27} \\
    \hline
\end{tabular}}}
\end{table*}

\begin{table*}[h]
\large
\centering
\setlength{\abovecaptionskip}{0pt}  
\setlength{\belowcaptionskip}{5pt}
\caption{3D detection scores per category on the SUN RGB-D, evaluated with mAP@0.25 IoU.}
\label{tab:25sun}
\renewcommand{\arraystretch}{1.5}{
\resizebox{0.75\textwidth}{!}{
\begin{tabular}{c|c|c|c|c|c|c|c|c|c|c|c}
    \toprule[1.5pt]
    &bathtub &bed &bookshelf &chair &desk &dresser &nightstand &sofa &table &toilet &mAP \\
    \hline
    VoteNet~\cite{qi2020imvotenet} &75.50 &85.60 &31.90 &77.40 &24.80 &27.90 &58.60 &67.40 &51.10 &90.50 &59.10 \\
    MLCVNet~\cite{xie2020mlcvnet} & 79.20 & 85.80 & 31.90 & 75.80 & 26.50 & 31.30 & 61.50 & 66.30 & 50.40 & 89.10 & 59.80\\ 
    H3DNet~\cite{zhang2020h3dnet} & 73.80 & 85.60 & 31.00 & 76.70 & 29.60 & 33.40 & 65.50 & 66.50 & 50.80 & 88.20 & 60.10\\
    BRNet~\cite{cheng2021back} &76.20 &86.90 &29.70 &77.40 &29.60 &35.90 &65.90 &66.40 &51.80 &91.30 &61.10 \\
    HGNet~\cite{chen2020hierarchical} & 78.00 & 84.50 & \textbf{35.70} & 75.20 & 34.30 & 37.60 & 61.70 & 65.70 & 51.60 & 91.10 & 61.60\\
    Group-free~\cite{liu2021group} &80.00 &87.80 &32.50 &79.40 &32.60 &36.00 &66.70 &70.00 &53.80 &91.10 &63.00 \\
    FCAF3D~\cite{rukhovich2021fcaf3d} &79.00 &88.30 &33.00 &81.10 &34.00 &40.10 &71.90 &69.70 &53.00 &91.30 &64.20 \\
    \hline
    Ours &\textbf{81.37} &\textbf{90.81} &32.64 &\textbf{82.97} &\textbf{39.19} &\textbf{42.74} &\textbf{73.49} &\textbf{72.22} &\textbf{59.64} &\textbf{92.91} &\textbf{66.80} \\
    \hline
\end{tabular}}}
\end{table*}

\begin{table*}[h]
\large
\centering
\setlength{\abovecaptionskip}{0pt}  
\setlength{\belowcaptionskip}{5pt}
\caption{3D detection scores per category on the SUN RGB-D, evaluated with mAP@0.50 IoU.}
\label{tab:50sun}
\renewcommand{\arraystretch}{1.5}{
\resizebox{0.75\textwidth}{!}{
\begin{tabular}{c|c|c|c|c|c|c|c|c|c|c|c}
    \toprule[1.5pt]
     &bathtub &bed &bookshelf &chair &desk &dresser &nightstand &sofa &table &toilet &mAP \\
    \hline
    VoteNet~\cite{qi2020imvotenet} &45.40 &53.40 &6.80 &56.50 &5.90 &12.00 &38.60 &49.10 &21.30 &68.50 &35.80 \\
    H3DNet~\cite{zhang2020h3dnet} & 47.60 & 52.90 & 8.60 & 60.10 & 8.40 & 20.60 & 45.60 & 50.40 & 27.10 & 69.10 & 39.00\\
    BRNet~\cite{cheng2021back} & 55.50 & 63.80 & 9.30 & 61.60 & 10.00 & 27.30 & 53.20 & 56.70 & 28.60 & 70.90 & 43.70\\
    Group-free~\cite{liu2021group} &64.00 &67.10 &12.40 &62.60 &14.50 &21.90 &49.80 &58.20 &29.20 &72.20 &45.20\\
    FCAF3D~\cite{rukhovich2021fcaf3d} &66.20 &\textbf{69.80}&11.60 &68.80 &14.80 &30.10 &59.80 &58.20 &35.50 &\textbf{74.50} &48.90 \\
    \hline
    Ours &\textbf{68.55} &67.44 &\textbf{13.82} &\textbf{70.84} &\textbf{17.28} &\textbf{30.92} &\textbf{59.91} &\textbf{61.27} &\textbf{39.22} & 72.73 &\textbf{50.20} \\
    \hline
\end{tabular}}}
\end{table*}

\section{More Results}
\label{sec:exp}
\subsection{Per-class Evaluation}
\label{sec:per-class}
We evaluate per-category on ScanNet V2 and SUN RGB-D under different IoU thresholds. Table \ref{tab:25scan}, \ref{tab:50scan} report the results on 18 classes of ScanNet V2 with 0.25 and 0.5 box IoU thresholds respectively. Table \ref{tab:25sun}, \ref{tab:50sun} show the results on 10 classes of SUN RGB-D with 0.25 and 0.5 box IoU thresholds. Our approach outperforms the baseline VoteNet~\cite{qi2020imvotenet} and previous state-of-the-art method FCAF3D~\cite{rukhovich2021fcaf3d} significantly in almost every category. Notably, our model significantly performs better than prior works on tiny classes (e.g., picture: +10.80 and +13.48 better than the SOTA on ScanNet V2), which demonstrates the effectiveness of our local grouping strategy.

\subsection{Latency and Memory Analysis.}
\label{sec:runtime}
We also report the latency and memory usage of our CAGroup3D on ScanNet V2. For a fair comparison, we re-measure all the methods on the same workstation (Single NVIDIA RTX 3090 GPU card, 256G RAM, and Xeon(R) E5-2638 v3) and enviroment (Unbuntu-16.04, Python 3.7, Cuda-11.1 and Pytorch-1.8.1). The official code of other methods is used for evaluation. Table \ref{tab:runtime} shows that our method achieves better performance with a competitive speed. The time cost of CAGroup3D is mainly on class-aware local grouping step, which iterates over all the
classes to generate high-quality 3D proposals. However, in our approach, we use semantic threshold to select a point subset for each category, which can significantly reduce the computation usage. To achieve faster running speed, we also present a light-weight version with the larger voxel size (0.04m). With this modification, our model can be faster and still maintain a high performance. In addition, we further add the inference time comparison between our RoI-Conv module and other alternatives in Table \ref{tab:roi_ab}. It can be seen that RoI-Conv pooling module is significantly more memory-and-time efficient than previous pooling operation. Hope it can be useful for the following two-stage methods.

\begin{table}[h]
  	\caption{Performance comparison of latency and runtime memory on ScanNet V2 dataset. All methods are tested on same workstation.}
  	\label{tab:runtime}
  	\footnotesize
  	\centering
  	\begin{tabular}{ccccccc}
    	\toprule
    	Method  & 1-stage & 2-stage & total latency & memory & mAP@0.25 & mAP@0.5\\
    	\hline
    	VoteNet~\cite{qi2020imvotenet}          & 101.0ms  & - & 101.0ms & 2,507MB  & 58.6 & 33.5   \\
    	Group-free~\cite{liu2021group}           & 153.1ms & -     & 153.1ms   & 3,678MB & 69.1 & 52.8    \\
    	FCAF3D~\cite{rukhovich2021fcaf3d}	   & 114.9ms & -        & 114.9ms   & 3,755MB  & 71.5 & 57.3    \\
    	Ours(light)	     & 111.6ms & 12.3ms     & 123.9ms    & 2,947MB  & 74.0 & 60.1 \\
    	Ours		     & 144.8ms & 34.5ms     & 179.3ms    & 3,544MB  & \textbf{75.1} & \textbf{61.3} \\
    \hline
  	\end{tabular}
\end{table}

\begin{table}[h]
  	\caption{Comparison with other RoI pooling approaches.}
  	\label{tab:roi_ab}
  	\small
  	\centering
   	\begin{tabular}{ccccc}
    	\toprule
    	RoI Method &mAP@0.25 & mAP@0.5 & memory & speed\\
    	\midrule
    	PointRCNN   & 73.65        & 57.83    & 8,054MB   & 62.9ms \\
    	Part-A$^2$        & 74.01        & 58.89   & 6,540MB  & 47.9ms \\
    	Ours-SA	      & 73.89        & 58.14   & 11,508MB   & 45.5ms  \\
    	Ours-SpConv		  & \textbf{74.50}        & \textbf{60.31}    & \textbf{2,468MB}  & \textbf{34.5ms} \\
    	\bottomrule
  	\end{tabular}
\end{table}

\subsection{More Ablation Studies}
\label{sec:more_ab}
\noindent\textbf{The effect of feature offsets.} We provide the ablative studies of the feature shifting operation in our class-aware grouping module on ScanNet V2. We can observe in Table \ref{tab:feat_shift} that feature shifting is slightly better than the variant of non-shift. As discussed in VoteNet, to generate more reliable object representations, a MLP is used to transform seeds’ features extracted from backbone to vote space, so that the grouped features can align with the voted points automatically.

\noindent\textbf{The effect of different loss weights.} As mentioned in \S \ref{sec:loss}, we simply set all the loss weights to 1.0 except for the bbox refinement $\beta_{rebox}$, which is adjusted to 0.5 for balancing the value of Stage-I box loss $\mathcal{L}_\text{box}$ and Stage-II refinement loss $\mathcal{L}_\text{rebox}$. Our method is not sensitive to loss weight and causes only minimal fluctuations (\textit{e.g.} less than 0.3) as shown in Table \ref{tab:ab_loss}.
\begin{table}[t]
\begin{minipage}[b]{0.48\linewidth}
  	\caption{Ablation study of feature shifting.}
  	\label{tab:feat_shift}
  	\small
  	\centering
   	\begin{tabular}{ccc}
    	\toprule
    	Feature Shifting &mAP@0.25 & mAP@0.5 \\
    	\midrule
    	       & 74.18        & 60.17   \\
        \checkmark  & \textbf{74.50}        & \textbf{60.31}   \\
    	\bottomrule
  	\end{tabular}
\end{minipage}
\hspace{12pt}
\begin{minipage}[b]{0.48\linewidth}
  	\caption{Ablation study of loss weight.}
  	\label{tab:ab_loss}
  	\small
  	\centering
   	\begin{tabular}{ccc}
    	\toprule
    	$\beta_{rebox}$ &mAP@0.25 & mAP@0.5 \\
    	\midrule
    	1.0 & 74.29        & 60.15 \\
        0.5  & \textbf{74.50}        & \textbf{60.31}   \\
    	\bottomrule
  	\end{tabular}
\end{minipage}
\end{table}

\noindent\textbf{More possible combinations of the imporatant modules.} In Table \ref{tab:combination}, we list more results of combining different modules mentioned in the main paper including Semantic Prediction, Diverse Local Group, RoI-Conv and BiResNet. It can be seen that our well-designed modules can still boost the performance with various combinations, which shows the robustness and effectiveness of our method.
\begin{table}[h]
  \caption{Effect of Semantic Prediction, Diverse Local Group, RoI-Conv and BiResNet.}
  \label{tab:combination}
  \centering
  \small
  \begin{tabular}{cccccc}
    \toprule
    Semantic Prediction & Diverse Local Group & BiResNet & RoI-Conv & mAP@0.25 & mAP@0.5\\
    \midrule
    		     	    &                     &            & \checkmark  & 69.10        & 57.62      \\
    \checkmark   		& \checkmark          &            & \checkmark  & 73.14            & 59.85     
        \\
                  		&                     & \checkmark & \checkmark  & 70.99        &  58.42     \\
    \checkmark   		& \checkmark          & \checkmark & \checkmark & \textbf{74.50}        & \textbf{60.31}      \\
	
    \bottomrule
  \end{tabular}
\end{table}

\subsection{Quantitative Results}
\label{sec:vis}
\begin{figure}[h]
  \centering
  \includegraphics[width=1.0\linewidth]{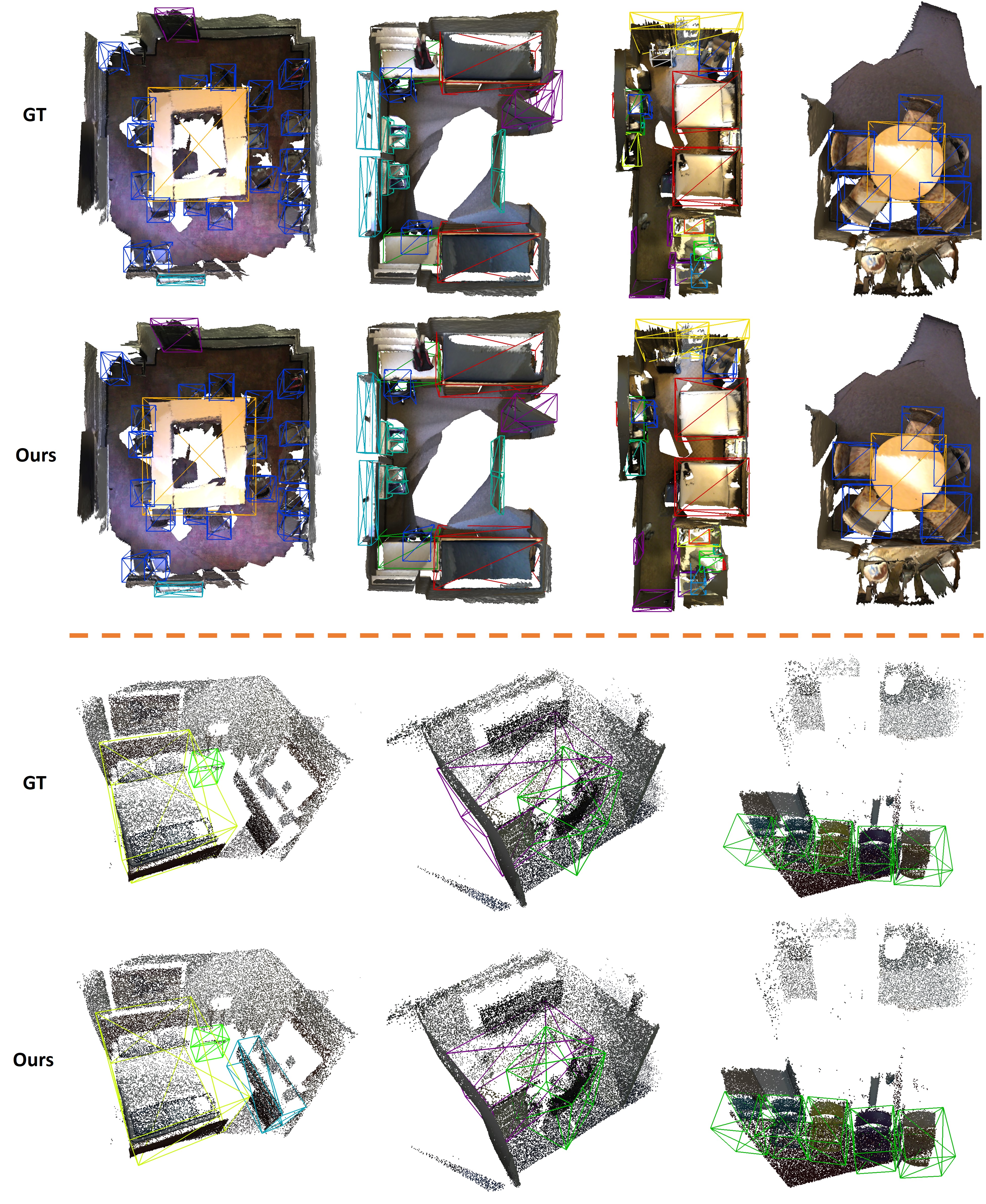}
  \caption{Qualitative results on ScanNet V2(top) and SUN RGB-D(down).}
  \label{fig:vis}
\end{figure}

We provide the visualization of our prediction bounding boxes on ScanNet V2 and SUN RGB-D datasets. Please see Figure \ref{fig:vis} for more qualitative results. Notably, our method can even accurately detect some miss-annotated objects in SUN RGB-D as in the bottom left of the figure.  

\end{document}